\documentclass[journal,comsoc]{IEEEtran}
\newcommand{\bl}{\textcolor{black}}
\usepackage{graphicx,epsfig,amssymb,amsmath,url,latexsym,stfloats,array,bm}
\usepackage[tight,footnotesize]{subfigure}
\usepackage{makecell}
\usepackage{mathrsfs}
\usepackage{amsthm}
\usepackage{longtable}
\usepackage{amsfonts}
\usepackage{caption}
\usepackage{graphicx}
\usepackage{multirow}
\usepackage{longtable}
\usepackage{float}
\usepackage{afterpage}
\usepackage{subeqnarray}
\usepackage{epstopdf}
\usepackage{empheq}
\usepackage{latexsym}
\usepackage{CJK}
\usepackage{color, soul}
\usepackage{framed}
\usepackage{lipsum}
\usepackage{color}
\usepackage{stfloats}
\usepackage{setspace}
\definecolor{shadecolor}{rgb}{1,0,0}

\usepackage{algorithmic}
\usepackage[linesnumbered,ruled]{algorithm2e}

\begin{document}
\title{Deep Reinforcement Learning for Delay-Oriented IoT Task Scheduling in Space-Air-Ground Integrated Network}
\author{\small Conghao~Zhou,~\IEEEmembership{\small Student~Member,~IEEE,}
            Wen~Wu,~\IEEEmembership{\small Member,~IEEE,}
            Hongli~He,~Peng Yang,~\IEEEmembership{\small Member,~IEEE,}\\
	        Feng~Lyu,~\IEEEmembership{\small Member,~IEEE,}
	        Nan~Cheng,~\IEEEmembership{\small Member,~IEEE,} and
	        Xuemin~(Sherman)~Shen,~\IEEEmembership{\small Fellow,~IEEE}

\thanks{C.~Zhou, W.~Wu and X.~Shen are with the Department of Electrical and Computer Engineering, University of Waterloo, Waterloo, ON, N2L 3G1, Canada (e-mail:\{c89zhou, w77wu, sshen\}@uwaterloo.ca).}
\thanks{H.~He is with the School of Information Engineering, Zhejiang University, Hangzhou, 310027, P.R. China (e-mail:~hongli\_he@zju.edu.cn).}
\thanks{P.~Yang is with the School of Electronic Information and Communications, Huazhong University of Science and Technology, Wuhan, 430074, P.R. China (e-mail:~yangpeng@hust.edu.cn).}
\thanks{F.~Lyu is with the School of Computer Science and Engineering, Central South University, Changsha, 410083, P.R. China (e-mail:~fenglyu@csu.edu.cn).}
\thanks{N.~Cheng is with Key State Lab. of ISN, and the School of Telecommunications Engineering, Xidian University, Xi'an, 710071, P.R.~China (e-mail:~nancheng@xidian.edu.cn). 
}

\thanks{Part of this work has been presented at IEEE GLOBECOM 2019 \cite{chau2019delay}.}
}	
\maketitle

\begin{abstract}
In this paper, we investigate a computing task scheduling problem in space-air-ground integrated network (SAGIN) for delay-oriented Internet of Things (IoT) services. In the considered scenario, an unmanned aerial vehicle (UAV) collects computing tasks from IoT devices and then makes online offloading decisions, in which the tasks can be processed at the UAV or offloaded to the nearby base station or the remote satellite. Our objective is to design a task scheduling policy that minimizes offloading and computing delay of all tasks given the UAV energy capacity constraint. To this end, we first formulate the online scheduling problem as an energy-constrained Markov decision process (MDP). Then, considering the task arrival dynamics, we develop a novel deep risk-sensitive reinforcement learning algorithm. Specifically, the algorithm evaluates the risk, which measures the energy consumption that exceeds the constraint, for each state and searches the optimal parameter weighing the minimization of delay and risk while learning the optimal policy. Extensive simulation results demonstrate that the proposed algorithm can reduce the task processing delay by up to 30$\%$ compared to probabilistic configuration methods while satisfying the UAV energy capacity constraint.

\end{abstract}

\begin{IEEEkeywords}
Space-air-ground integrated network, IoT, edge computing, reinforcement learning, constrained MDP.
\end{IEEEkeywords}

\section{Introduction}
\bl{Equipped with advanced embedded monitoring and data collection technologies, Internet of Things (IoT) devices, such as high definition cameras, object detectors, and meteorological sensors, play vital roles in a myriad of applications and services~\cite{wang2020optimal}. Specifically, IoT devices can be deployed to monitor and sense the environment, offering new opportunities for industrial automation, intelligent transportation management, etc. There are two typical applications of delay-oriented IoT services: intelligent urban transportation management and automated surface mining in suburban areas. For intelligent transportation management, on-board cameras and road-side sensors can reliably detect incidents, such as traffic signal violations, stopped vehicles, and on-road pedestrians. By leveraging deep learning-based image processing techniques, vehicle and pedestrian behaviors can be predicted to prevent potential traffic accidents in advance~\cite{shen2020ai}. Rapidly processing the collected image can save more time in reacting to the complicated transportation scenarios, which enhances the road safety by preventing the transportation emergency. For automated surface mining, a large number of cameras and visual sensors are deployed in the active areas of the drill rigs to assess rock composition and collect environment information (e.g., humidity and temperature). The analytics results of input image/video from these IoT devices can help achieve automated drilling control~\cite{wang2020convergence}. In this case, lower delay of image/video analytic can enable more accurate automated surface mining control. Generally, such IoT services are delay-oriented which should be processed rapidly to adapt to highly dynamic input.}

\bl{To support the aforementioned services, ubiquitous delay-oriented computing tasks become prevailing on IoT devices, resulting in a surging demand for computing capability~\cite{yang2019edge}. Due to the limited computing capability of IoT devices, executing these delay-oriented tasks locally, such as on-camera image/video processing, can inflict unacceptable service delay and be detrimental to the service lifespan of IoT devices~\cite{zhou2019online}.}
Edge computing has been proposed as a de-facto paradigm to support computation-intensive IoT services. Within this paradigm, IoT devices can offload computing tasks to nearby terrestrial base stations (BSs), which can not only reduce the latency of task execution, but also save the power consumption of IoT devices~\cite{zhou2019edge}. 
However, purely relying on offloading to terrestrial BSs is hard to guarantee the performance of IoT edge computing robustly. 
On the one hand, the IoT devices are usually power constrained, which cannot support long-distance transmission for task offloading, especially when the BSs are sparsely deployed or unavailable nearby (e.g., automated mining applications)~\cite{8672604}. On the other hand, the physical computing resources on BSs are scarce and somewhat insufficient, but the IoT computing tasks arrive dynamically with possible bursty conditions (e.g., intelligent transportation applications), which can result in computing resource shortage and deteriorate delay performance~\cite{huo2019secure}~\cite{reliable-V2V}.

As a remedy to these limitations, satellites and unmanned aerial vehicles (UAVs) are considered as promising complements to enhance the terrestrial network. 
For satellites, many research and industrial efforts have been devoted to the commercialization of the low earth orbit (LEO) satellite constellation, such as SpaceX and OneWeb~\cite{buchen2015small}, which can provide ubiquitous services with acceptable propagation delay (e.g., about 6.44\,ms)~\cite{3GPP2019,di2019ultra}. 
For UAVs with flexible deployment and agile management, they have been widely utilized in military and civil applications to provide on-demand communication and computing resources~\cite{wu2018joint}.
Besides, the 3rd Generation Partnership Project (3GPP) is also investigating on non-terrestrial networks and specifying novel architectures to complement terrestrial cellular networks~\cite{3GPP2019}.
\bl{Since satellite, UAV, and BS can complement each other, the integration of them, namely the space-air-ground integrated network (SAGIN), is proposed as a promising next-generation wireless network to serve the massive IoT devices with delay-oriented service requirements~\cite{8672604},~\cite{zhang2018air}.}

\bl{In this paper, considering the low transmit power and short-distance communication range of IoT devices, we propose a \underline{d}elay-\underline{o}rientated IoT \underline{t}ask \underline{s}cheduling (DOTS) scheme in SAGIN to process computing tasks in real time.}
We adopt a UAV (installed with dedicated IoT communication interface such as LoRa and NB-IoT~\cite{8030322},~\cite{7422054}) as the ``flying scheduler'' to communicate with IoT devices and collect their computing tasks. 
As the UAV can move sufficiently close to IoT devices, the distance between IoT devices and the UAV can be significantly reduced, which not only saves the IoT devices' power consumption and prolongs the service lifespan, but also guarantees the transmission reliability~\cite{kato2019optimizing}.
Then, the UAV makes task scheduling decisions in real time, i.e., processing locally, offloading to a nearby BS or the remote LEO satellite constellation.\footnote{Note that the UAV can be installed with two communication interfaces, one for cellular BSs and the other for the LEO satellite constellation in SAGIN~\cite{8672604,li2018investigation}.} 
Particularly, the UAV needs to offload tasks as soon as possible when it serves an excessive number of IoT computing tasks, due to the limited computing capability~\cite{zhou2018sagecell}.
In addition, the UAV should make decisions in real time to keep the pace of dynamic link conditions and computing task arrival. 
Therefore, how to obtain an efficient scheduling policy of processing IoT computing tasks at appropriate SGAIN components is a crucial issue, which is quite challenging due to the following three reasons.
First, with a large number of IoT devices, task arrivals are dynamic and may be bursty, and even unknown \emph{a priori}, which poses a real-time requirement for the scheduling policy. 
Second, UAV, BSs, and LEO satellites have differentiated features in terms of communication and computing capability. As a result, the scheduling policy should select appropriate SAGIN components for task processing in accordance with their features.
Third, in the scheduling policy, both the current energy consumption and the energy reservation for future arrived tasks should be considered. The UAV needs to comply with the UAV energy capacity by making sequential task scheduling decisions.

To tackle the above challenges, we formulate the online scheduling problem as a constrained Markov decision process (CMDP) to minimize the time-averaged task processing delay while taking the UAV energy capacity (consumed by communication and computing) into consideration. 
Inspired by the advantage of reinforcement learning (RL) methods in tackling the uncertainty and dynamics, we design a novel deep risk sensitive RL algorithm to deal with the formulated CMDP problem.
The core idea is to define a risk function to capture whether the UAV energy capacity constraint is violated. Thus, satisfying the
constraint is transformed into minimizing the risk. Afterward, we replace the typical Q-value function by the sum of two Q-value functions. The former Q-value function evaluates the long-term delay for different state-action pairs, and the latter accounts for the long-term risk. Based on the designed Q-value function, the scheduling policy can be learned by leveraging RL methods. Meanwhile, instead of constructing a space-costly Q-value table caused by the high dimensional state representation, we leverage the parameterized deep neural network (DNN) to approximate the Q-value function. In addition, we add a filter layer after fully connected layers to exclude unavailable actions at different states.
Extensive simulations are conducted, which show that the proposed deep RL-based DOTS scheme can achieve a lower time-average task processing delay while satisfying the UAV energy capacity constraint compared to that of benchmark schemes. The main contributions of this paper are three-fold: 
\begin{itemize}
\item We propose a computing task scheduling scheme named DOTS for delay-oriented IoT services in SAGIN, where a UAV flies along a trajectory to collect computing tasks and make real-time scheduling decisions. 
\item We formulate an integer non-linear optimization problem with uncertainty to minimize the time-averaged task processing delay under the UAV energy capacity constraint. As the UAV location and task backlog evolve in an ergodic way, we reformulate the online IoT task scheduling problem as a CMDP.
\item We design a novel deep risk-sensitive RL algorithm to address the CMDP problem, where a risk function is defined to indicate whether the UAV energy consumption violates the constraint. Besides, we leverage DNNs to implement the proposed deep RL-based algorithm in the DOTS scheme.
\end{itemize}

The remainder of this paper is organized as follows.
Section II presents the related work.
We describe the SAGIN architecture and computing task scheduling models in Section III.
In Section IV, we provide the problem formulation.
We design the DOTS scheme to make the online scheduling decision in Section V.
Section VI presents the simulation results of DOTS, followed by the conclusion and the future work in Section VII.

\section{Related Work}
SAGIN is envisioned as a promising architecture to complement the terrestrial network for the next-generation wireless network. 
To guarantee service requirements in dynamic and heterogeneous SAGIN, a cost-effective scheme for joint service placement and routing is proposed in \cite{varasteh2019mobility}.
To accommodate diverse services, resources of the satellite, aerial, and terrestrial components have been sliced, and a hierarchical resource management scheme is proposed to put available resources into a common and dynamic resource pool \cite{zhang2017software}. 
To meet the emerging computation-intensive IoT applications with diverse QoS requirements, an air-ground integrated mobile edge network is presented to realize mobile edge computing \cite{cheng2018air}. In~\cite{8672604}, to address uncertain channel conditions in remote areas, an RL-based scheduling scheme is proposed for the virtual machine assignment and task offloading in SAGIN. However, accommodating IoT computing task scheduling in SAGIN still faces significant challenges since the computing task arrival from IoT devices is highly dynamic and random, and the management for both communication and computing resources is complicated.

Although the research on IoT computing task scheduling in SAGIN is at its initial stage, applying task scheduling for IoT devices in other scenarios has been exploited extensively. To solve the joint problem of partial offloading scheduling and resource allocation for mobile edge computing systems with multiple independent tasks, a two-level alternation method is proposed based on the Lagrangian dual decomposition \cite{kuang2019partial}. To address the multi-user computation offloading problem for mobile-edge cloud computing in a multi-channel wireless interference environment, a distributed computation offloading algorithm is proposed based on a Nash equilibrium \cite{chen2016efficient}. However, it is difficult for an optimization-based algorithm to adapt to the dynamic task arrival scenario since a fixed task number is required. Considering the stochastic task generation, Lyapunov optimization is leveraged in task scheduling schemes. 
Besides, an asymptotically optimal scheduling scheme is also proposed with partial knowledge in mobile edge computing scenarios by leveraging the Lyapunov drift~\cite{lyu2017optimal}.
In order to minimize the delay due to both radio access and computation, a user-centric energy-aware mobility management scheme is proposed based on Lyapunov functions and multi-armed bandit theories \cite{sun2017emm}. The Lyapunov-drift-based techniques can schedule tasks to keep the task queue stable based on the current queue backlog. However, the optimality cannot be guaranteed since the information of future status (e.g., future task arrival) is lacking.

Preliminary results of this work have been presented \cite{chau2019delay}, in which the task arrival pattern is assumed to be known to the UAV. In practice, this information may be difficult to be obtained. To accommodate to dynamic task arrival, we propose an IoT task scheduling scheme in SAGIN relying on deep risk-sensitive RL to minimize the time-averaged task processing delay while considering the UAV energy capacity.

\section{System Model}
In this section, we first introduce the proposed DOTS scheme in SAGIN architecture, and then describe the computing, communication, and energy consumption models for IoT task offloading.

\subsection{The SAGIN Architecture and the DOTS Scheme}
\begin{figure}[t]
	\centering
  	\includegraphics[width = 1\columnwidth]{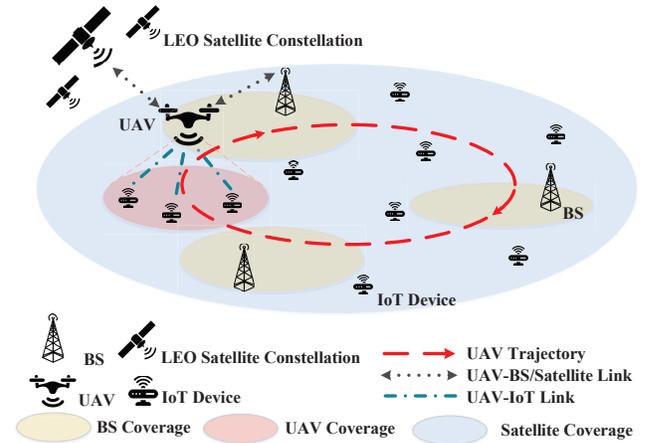}
  	\caption{\small The network model.}\label{system}
\end{figure}
\bl{As shown in Fig.~\ref{system}, the UAV flies along a trajectory to collect delay-oriented computing tasks from IoT devices.\footnote{The UAV trajectory is assumed to be planned in advance since the UAV trajectory design has been well studied in many previous works~\cite{cheng2018uav,zeng2017energy}.} As the rotary-wing UAV can hover in the air, and fly with a low height sufficiently close to IoT devices, we adopt the rotary-wing UAV to collect the computing tasks~\cite{zeng2017energy}. Taking the computing functionality of the UAV~\cite{wu2018joint}, BSs~\cite{chen2016efficient}, and LEO satellites~\cite{8672604} into account in the SAGIN, the UAV can schedule computing tasks on three different destination network components, i.e., processing tasks on the UAV locally, offloading to the nearby BS, or offloading the LEO satellite constellation. Let indexes $1, 2, \ldots , N$, and $0$ denote the LEO satellite constellation and the BSs, respectively. Then, the set of the network components that computing tasks can be offloaded to (i.e., $N$ BSs and the LEO satellite constellation) is denoted by $\mathcal N$ = $\{0, 1, 2, \ldots , N \}$. Due to the UAV's limited on-board battery capacity, the computing capability at the UAV is limited \cite{wu2018joint}. The UAV cannot process all computing tasks alone, and thus some computing tasks can be offloaded to BSs or the LEO satellite constellation. BSs and the LEO satellite constellation have different characteristics. The BS has high computing capacity, while its coverage area is limited. The LEO satellite constellation can always cover the area and act as a complementary offloading solution for terrestrial networks, while the propagation delay of the UAV-satellite link cannot be neglected. Therefore, computing tasks should be scheduled appropriately to different destination network components in SAGIN to reduce the service delay.}

\begin{figure*}
	\centering
  	\includegraphics[width = 1.8\columnwidth]{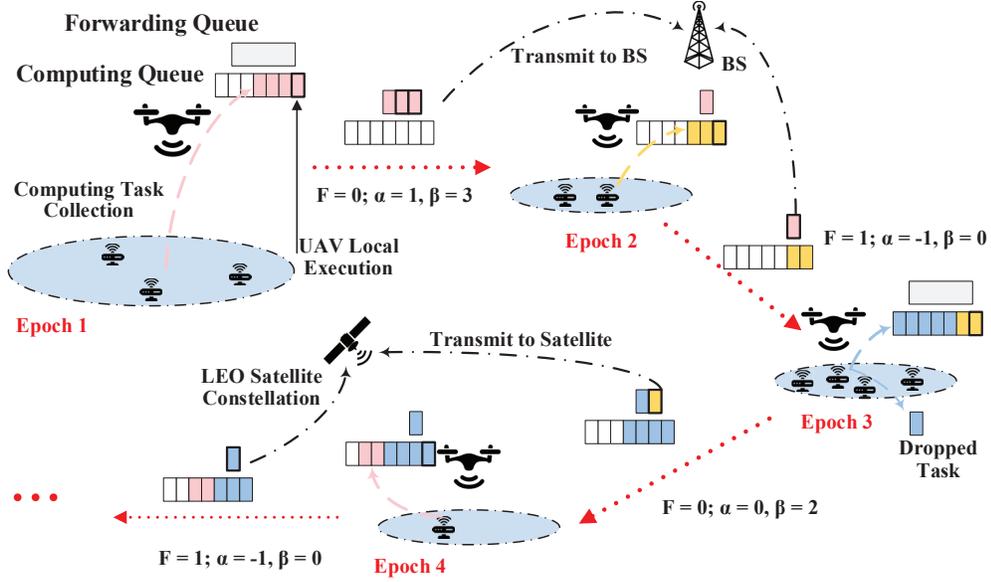}
  	\caption{\small An illustration of the DOTS scheme in SAGIN, where different colors of tasks are used to distinguish the collection in different epochs.}\label{flow}

\end{figure*}

\bl{We adopt the discrete epoch-based system with an equal time duration of $\tau$ in each epoch. In epoch $t$, the location of the deployed UAV is denoted by $l_t$. As the UAV flies along the trajectory, the set of available offloading destination network components also varies at different locations, which is denoted by $\mathcal{L}_t \subseteq \mathcal{N}$. Supposing that multiple computing tasks can be offloaded from the UAV in each epoch, only one offloading destination (i.e., a BS or the satellite) can be chosen.}
In summary, the UAV collects and schedules IoT computing tasks according to the following steps in each epoch:
\begin{itemize}
\item[1)] \bl{The UAV collects tasks from IoT devices and locally processes their tasks within the computing queue. The collected tasks that have not been processed or offloaded will wait in the computing queue at the UAV.}
\item[2)] \bl{The UAV can offload a certain number of computing tasks from the computing queue to a BS or the satellite. The offloaded tasks that have not been forwarded will wait in the forwarding queue at the UAV.}
\item[3)] Newly arrived tasks from IoT devices are stored in the computing queue at the UAV. Once the computing queue is full, newly arrived tasks will be dropped.
\item[4)] The UAV flies to the next location along the predefined trajectory, and continues to collect computing tasks. 
\end{itemize}
As shown in Fig. \ref{flow}, an exemplary work flow of the DOTS scheme in SAGIN is illustrated. In epoch~1, four tasks are collected, one of which is processed locally at the UAV, and three of which are offloaded to BS and moved into the forwarding queue. In epoch~2, the UAV cannot move new tasks into the forwarding queue due to the uncompleted task forwarding. Only one task is processed locally at the UAV, and all tasks in the forwarding queue are transmitted. In epoch~3, two tasks are offloaded to the satellite and moved into the forwarding queue. In epoch~4, all tasks can only be executed locally at the UAV. The details of the scheme are introduced in the following subsections.

\subsection{Computing Model}
In general, we adopt a tuple $\left(\phi,\gamma\right)$ to model a computing task~\cite{8672604}. Here, \bl{$\phi$ represents the input data size (in bits) of a computing task, and $\gamma$ (in central processing unit (CPU) cycles per bit) indicates the computing workload of the task, i.e., how many CPU cycles are required to process one bit input data.\footnote{In practice, the computing workload is measured via conducting the same computing task with the same software of experimental platform in multiple times~\cite{hennessy2011computer}.}} \bl{Note that task uploading is the key point of scheduling policy at the UAV in the considered scenario, and the downloading of the computing result can be ignored in this work.\footnote{Generally, the results of computing tasks cannot be immediately fed back to IoT devices by the same UAV due to the mobility. In practical system, many UAVs can be deployed along different pre-defined trajectories, and the result of computing tasks can be relayed via UAV-UAV links \cite{wang2019adaptive}. Therefore, IoT devices can receive results as long as they are covered by UAVs.} For instance, IoT devices upload images for analysis and download text messages as the output, and the uploaded data size is much larger than that of downloaded data~\cite{yang2019edge}.
As the UAV can offload tasks to either the nearby BS or the remote LEO satellite, or execute tasks locally, the corresponding computing delay is analyzed in the next.}

\subsubsection{Task Offloading}
\bl{Denote the task offloading decision by $\alpha_t$ in epoch $t$, i.e., the offloading destination network components in epoch $t$. The UAV offloads the tasks to the satellite when $\alpha_t = 0$, or offloads tasks to BS $n$ when $\alpha_t = n, \forall n \not= 0$. Denote by $\beta_t \leq \beta^\text{max}, \beta_t \in \mathbb{N}$ the number of offloaded tasks in epoch $t$, where $\beta^\text{max}$ is the maximal number of tasks that can be forwarded by the UAV in each epoch. 
Meanwhile, due to the occupation of the communication interface, we assume that new tasks cannot be forwarded if the offloading process of the last task is not completed. Let binary variable $F_t$ indicate whether collected IoT tasks on the UAV can be offloaded or not.}
Fig.~\ref{flow} illustrates an example of the task forwarding. When $F_t = 0$, the UAV can offload tasks in epoch $t$ since the channel is not occupied (i.e., $\alpha_t \in \mathcal{L}_t, \beta_t \leq \beta^\text{max}$). $F_t = 1$ represents that the UAV cannot offload new tasks since a certain number of tasks are waiting to be transmitted in the forwarding queue (i.e., $\alpha_t = -1, \beta_t = 0$). 

Denote by the computing capabilities (in CPU cycles per second) of BS $n$ and the satellite by $f_n, n \not= 0$ and $f_0$, respectively. The computing delay of all $\beta_t$ tasks at offloading destination network component $n$ is given by:
\begin{equation}\label{eq1}
d_1(\alpha_t, \beta_t) = \frac{\beta_t \phi \gamma}{f_{\alpha_t}}, \,\, \alpha_t \in \mathcal{L}_t,
\end{equation}
where $f_{\alpha_t}$ represents the computing capability of offloading destination network component $\alpha_t$.   

\subsubsection{Local Processing} 
\bl{Since the computing capability of the UAV is limited, the collected tasks may not be processed locally or offloaded completely at the UAV within an epoch. We assume that the remaining tasks wait to be scheduled in the computing queue at the UAV. As a result, the delay of processing task locally at UAV includes two parts, i.e., local computing delay and queuing delay. To model the computing queue, we first denote the unaccomplished task backlog at the beginning of epoch $t$ by $H_t \in \left[0, \rho \right]$, where $\rho$ is the maximum length of the computing queue. Then, given unaccomplished task backlog $H_t$ and the number of offloaded tasks $\beta_t$, the number of queuing tasks $O_t$ in epoch $t$ within the computing queue is given by:}
\begin{equation}\label{eq3}
    O_t =\max \left\{H_t - \lfloor \frac{f_\text{U} \tau}{\phi \gamma} \rfloor - \beta_t, 0 \right\},
\end{equation}
where $f_\text{U}$ is the computing capability (in CPU cycles per second) of the UAV, and $ \lfloor f_\text{U} \tau / \phi \gamma \rfloor$ is the greatest integer less than the number of tasks executed by the UAV in epoch $t$. Given the number of newly collected tasks $M_t$ from the IoT devices, the unaccomplished task backlog $H_{t+1}$ can be updated at the end of epoch $t$ as follows:
\begin{equation}\label{eq5}
    H_{t+1} =\min \left\{O_t + M_t,\, \rho \right\},
\end{equation}
where $\min \{ \cdot \}$ is the function to return the smallest value. \bl{Then, the delay of local task execution at the UAV can be calculated as the following equation:
\begin{equation}\label{eq2}
    d_2 (\alpha_t, \beta_t) =  \frac{ \min \left\{\lfloor \frac{f_\text{U} \tau}{\phi \gamma} \rfloor, H_t \right\} \phi \gamma}{f_\text{U}} + O_t \tau,
\end{equation}
where $\min \{\lfloor \frac{f_\text{U} \tau}{\phi \gamma} \rfloor, H_t \} \phi \gamma / f_\text{U}$ is the local computing delay within each epoch, and $O_t \tau$ is the queuing delay of all $O_t$ tasks waiting in the computing queue.
}
\subsection{Communication Model}
\bl{We suppose two communication interfaces are equipped in this work~\cite{zhang2014dynamic}, i.e., one for LEO satellites, and the other for BSs. Each of them uses different spectrum bands, which leads to no interference between BSs and the satellite~\cite{vatalaro1995analysis}. In the following, the transmission delay of offloading tasks to the satellite and the BSs are discussed in detail.\footnote{Considering the flexibility of the UAV, it can fly sufficiently close to the IoT devices such that the condition of UAV-IoT links is Line-of-sight (LoS). Since the LoS communications can achieve high data rate~\cite{wu2018joint, hosseini2018uav}, the transmission delay of UAV-IoT links is neglected.}}

\subsubsection{Offload to Satellite}
\bl{Currently, the wireless communications between an LEO satellite and terrestrial users are enabled by Ka or Ku frequency band, the channel condition of which is mainly impacted by the communication distance and the rain attenuation (rain fading) \cite{vatalaro1995analysis}. Supposing the meteorological environment remains stationary during the IoT task collection, the channel gain of the UAV-satellite link is mainly determined by the distance between the UAV and the satellite. Generally, the moving distance of the UAV (e.g., the maximum flight distance of the UAV is about 2\,km) is much shorter than the altitude of the satellite (e.g., the LEO satellites are with an altitude of 200\,km to 2,000\,km), which results in the negligible variation of the distance between the UAV and the satellite~\cite{8672604}. Therefore, the channel gain $h$ of the UAV-satellite link can be assumed to be the same with the location of UAV.
Then, the data rate of the UAV-satellite link in epoch $t$ denoted by $r_{\alpha_t}$ is given by:}
\begin{equation}\label{eq6}
    r_{\alpha_t}= W_\text{S}\log _2\left(1+\frac{P_\text{S} \cdot |h|^2}{\sigma_\text{S}^2}\right), \;\;\;\;\alpha_t = 0,
\end{equation}
where $W_\text{S}$ is the channel bandwidth of the UAV-satellite link, $P_\text{S}$ is the transmission power of UAV-satellite link, and $\sigma_\text{S}^2$ indicates the power of noise.
Due to the long distance between the LEO satellite and the UAV, the propagation delay cannot be ignored, which is denoted by $d_\text{S}$.
Thus, given offloading decision $\alpha_t$ and offloaded task number $\beta_t$, transmission delay of offloading tasks to the satellite can be calculated as following equation:
\begin{equation}\label{eq7}
d_3(\alpha_t, \beta_t) = \frac{\beta_t \phi}{r_{\alpha_t}} + d_\text{S}, \;\;\;\;\alpha_t = 0.
\end{equation}

\subsubsection{Offload to BS}
Denote by $K_{\alpha_t}, \alpha_t \neq 0$ the duration that UAV will stay in the coverage of BS $n$ since epoch $t$. 
\bl{As the UAV needs to guarantee that the forwarding process of all $\beta_t$ tasks can be completed before the UAV flies out of the BS's coverage, the number of forwarded tasks $\beta_t$ satisfies the following constraint:}
\begin{equation}\label{cons1}
	\arg \min_{k}\left(\sum_{i=t}^{t+k}{r_{\alpha_i} \tau} \geq \beta_t \phi \right) \leq K_{\alpha_t},\;\;\;\; \alpha_t \in \mathcal{L}_t, \alpha_t \neq 0,
\end{equation}
\bl{which means that the transmission time of $\beta_t$ tasks is shorter than the duration that the UAV stays in the BS's coverage. Notice that duration $K_{\alpha_t}$ can be known \emph{a priori} for the deployed UAV as it depends on the BSs' location and the UAV trajectory \cite{8672604}.}

Given the pathloss of the UAV-BS link $PL$, data rate $r_{\alpha_t}$ of the UAV-BS $n$ link can be calculated as
\begin{equation}\label{eq9}
    r_{\alpha_t} = W_\text{B}\log _2\left(1+\frac{P_\text{B} \cdot 10^{\frac{PL}{10}}}{\sigma_\text{B}^2}\right), \;\;\;\;\alpha_t \neq 0,
\end{equation}
where $W_\text{B}$ indicates the channel bandwidth of UAV-BS link, $P_\text{B}$ represents the transmission power of from the UAV to a BS, and $\sigma_\text{B}^2$ indicates the power of the background noise. 
Denote by $d_3$ the transmission delay of offloading tasks to the BS, which is given by:
\begin{equation}\label{eq10}
d_3(\alpha_t, \beta_t) = \frac{\beta_t \phi}{r_{\alpha_t}}, \;\;\;\; \alpha_t \in \mathcal{L}_t, \alpha_t \neq 0,
\end{equation}
where $\alpha_t$ and $\beta_t$ represent offloading destination and offloaded task number, respectively.

\subsection{Energy Consumption Model}
Generally, UAV energy consumption includes propulsion energy, communication-related energy, and computing-related energy. Since UAV propulsion energy is mainly depends on different trajectories and aircraft parameters, it can be considered as a constant in our work \cite{zeng2017energy}. 
\bl{Thus, we aim to guarantee the remaining components of energy consumption, i.e., computing-related and communication-related energy, do not exceed the UAV energy capacity.}
\bl{Denote by $e_\text{o}$ the communication-related energy caused by the transmission of tasks, which can be calculated as follows:}
\begin{equation}
e_\text{o}(\alpha_t, \beta_t)= \left\{
\begin{aligned}
&P_\text{S}  d_4(\alpha_t, \beta_t), &&\; \alpha_t = 0\\
&P_\text{B}  d_4(\alpha_t, \beta_t), &&\; \alpha_t \in \mathcal{L}_t, \alpha_t \neq 0.
\end{aligned}
\right.
\end{equation}
\bl{Meanwhile, processing computing task on the UAV also consumes energy, which depends on the computing workload of the computing task and the computing capability of the UAV. Denoted by $e_\text{l}$ the computing-related energy, which can be expressed as follows:}
\begin{equation}
    e_\text{l}(\alpha_t, \beta_t) = \min \left\{H_t\phi \gamma, f_\text{U} \tau \right\} \cdot \xi \left(f_\text{U}\right)^2,
\end{equation}
where $\xi$ indicates the effective switched capacitance determined by the chip architecture \cite{8672604}. 
\bl{Denote by $E_t$ the cumulative energy consumption in epoch $t$. Given the communication-related and computing-related energy consumption, the cumulative energy consumption can be calculated as the following equation:}
\begin{equation}
E_{t} = E_{t-1} + e_\text{o}(\alpha_t, \beta_t) + e_\text{l}(\alpha_t, \beta_t).
\end{equation}
\bl{The cumulative energy consumption can be leveraged to evaluate whether the UAV satisfies the energy capacity or not. }

\section{Problem Formulation}
\bl{In our work, we aim to minimize the long-term delay of all computing tasks while satisfying the UAV energy consumption constraint. The total delay of all tasks in epoch $t$ can be calculated as follows:
\begin{equation}
D_t = \left\{
\begin{aligned}
&\frac{\beta_t \phi \gamma}{f_{\alpha_t}} + \frac{ \min \left\{\lfloor \frac{f_\text{U} \tau}{\phi \gamma} \rfloor, H_t \right\} \phi \gamma}{f_\text{U}} + O_t \tau + \frac{\beta_t \phi}{r_{\alpha_t}} + d_\text{S}, \,\,\,\, \alpha_t = 0\\
&\frac{\beta_t \phi \gamma}{f_{\alpha_t}} + \frac{ \min \left\{\lfloor \frac{f_\text{U} \tau}{\phi \gamma} \rfloor, H_t \right\} \phi \gamma}{f_\text{U}} + O_t \tau + \frac{\beta_t \phi}{r_{\alpha_t}} , \,\,\,\,  \alpha_t \neq 0,
\end{aligned}
\right.
\end{equation}
where both the computing delay and the transmission delay are included.} Let $\bm{\alpha} = \left\{\alpha_t, \forall t \right\}$ and $\bm{\beta} = \left\{\beta_t, \forall t \right\}$ denote the set of task offloading decisions and the number of offloaded tasks in each epoch, respectively. As link availability and task arrival are highly dynamic, we concentrate on minimizing the time-averaged delay of all tasks. The delay minimization problem can be formulated as follows:
\begin{subequations}\label{p1}
\begin{align}
\textrm{P1:} \,\, \min_{\left\{ \bm{\alpha}, \bm{\beta} \right\} } \,\,& \lim_{T \rightarrow \infty}{ \frac{1}{T}\sum_{t=1}^{T}{D_t } }\\
\textrm{s.t.} \,\,& (\ref{cons1}),\\
& \lim_{T \rightarrow \infty}{ \frac{1}{T}\sum_{t=1}^{T}{\left[ e_\text{o}(\alpha_t, \beta_t) + e_\text{l}(\alpha_t, \beta_t)\right]}} \leq \varepsilon,\\
& \alpha_t \leq N, \alpha_t \in \mathcal{L}_t,\\
& \beta_t \leq \beta_\text{max}, \beta_t  \in \mathbb{N},
\end{align}
\end{subequations}
\bl{where (\ref{p1}a) is the objective that minimizes the time-average delay of all collected tasks over $T$ epochs, and (\ref{p1}b) limits the offloading destinations and the number of offloading tasks. (\ref{p1}c) restricts the time-averaged energy consumption of the UAV where $\varepsilon$ is the UAV energy capacity. (\ref{p1}d) and (\ref{p1}e) constrain task offloading decisions and the numbers of offloaded tasks, respectively. Problem P1 is an integer nonlinear optimization problem with unknown number of newly collected tasks in each epoch, which is difficult to solve. Considering the UAV location and the backlog of unaccomplished task in the computing queue evolve in an ergodic way, we adopt the stationary decision to address this problem, which is time-invariant and only depends on the current system status.} Therefore, the problem can be reformulated as a Markov decision process (MDP) for a stationary decision which is the optimal in the ergodic system~\cite{puterman2014markov}.

We define a tuple $\mathcal M:=\left< \bm S, \bm A, \bm P, \bm C, \bm \Pi \right>$ to model the MDP, which is a sequential decision-making process.
Specifically, $\bm S$ represents the set of states. $\bm A$ is the set of actions. $\bm P:=\bm S\times \bm A\times \bm S\rightarrow \mathbb R$ is set of state transition probabilities. $\bm C:= \bm S\times \bm A\rightarrow \mathbb R$ indicates the cost function. $\bm \Pi$ is the policy that is a decision rule mapping from a state $\bm s \in \bm{S}$ to an action $\bm a \in \bm{A}$. Meanwhile, $C(\bm s, \bm a)$ is defined as the cost when the system stays in state $\bm s$ with adopting action $\bm a$. For the aforementioned problem, the states, actions, and cost in an MDP model are formulated as follows.
\subsubsection{State} In epoch $t$, a tuple denoted by $\bm s_t = (l_t, F_t, H_t, E_t), \bm s_t \in \bm S$ is used to describe the system state, where $l_t$, $F_t$, $H_t$, $E_t$ represent UAV location, the number of offloaded tasks in the forwarding queue, the unaccomplished task backlog in the computing queue and the cumulative energy consumption, respectively.
\subsubsection{Action} An action is made based on the current state, and the decision is denoted by a tuple $\bm a_t = (\alpha_t, \beta_t), \bm a_t \in \bm A$ in epoch $t$, where $\alpha_t$ is used to indicate offloading destination, and $\beta_t$ denotes the number of the offloaded tasks.
\subsubsection{State Transition} The state transition includes four components: the update of $l_t$, which only depends on the predefined UAV trajectory and the evolutions of $F_t$, $H_t$, $E_t$, which are discussed in the preceding section.

\subsubsection{Cost Function} 
\bl{Considering an intuitive policy that the UAV does not offload tasks and keep the queue full, and almost all newly arrived tasks will be dropped. In such case, although the cost (delay) can be minimized, an excessive number of dropped tasks lead to practical infeasibility.} To minimize the cost while avoiding the excessive task dropping, a penalty $\Lambda_t$ is introduced as follows:
\begin{equation}
	\Lambda_t = \lambda \max \left(M_t + O_t - \rho, 0\right),
\end{equation}
where $\max \left(M_t + O_t - \rho, 0\right)$ represents the excessive number of the newly collected tasks will be dropped, and $\lambda$ is a constant penalty weight. With the objective of minimizing long-term delay of all IoT tasks, the cost function can be defined as $C(\bm s_t, \bm a_t) = D_t + \Lambda_t$, where $\Lambda_t$ is the penalty to avoid excessive drop of computing tasks. 

\subsubsection{Policy} Denote by $\bm \pi$ the stationary policy, which means that state $\bm s_t$ is assigned with action $\bm a_t$ and this action will be chosen whenever the system stays in this state.

Therefore, MDP based delay-oriented tasks scheduling problem can be formulated as follows:
\begin{subequations}\label{p2}
\begin{align}
\textrm{P2:} \,\, \min_{\bm \pi} \,\,&  \lim_{T \rightarrow \infty} \mathbb E \left[\frac{1}{T} \sum_{t = 1}^{T}{C_t(\bm s_t, \bm a_t)} \bigg| \bm \pi \right]\\
\textrm{s.t.} \,\,& (\ref{p1}\text{b}), (\ref{p1}\text{d}), (\ref{p1}\text{e})\\
& \lim_{T \rightarrow \infty}  \mathbb E \left[\frac{E_t}{T} \bigg| \bm \pi \right] \leqslant \varepsilon,
\end{align}
\end{subequations}
\bl{where (\ref{p2}a) represents the expected average cost and expected energy consumption. Problem P1 is transformed into problem P2 to find the optimal policy $\bm \pi$ with respect to a cost $C_t(\bm s_t, \bm a_t)$ for choosing action $\bm a$ at state $\bm s$, which minimizes the expected average cost. 
Above problem P2 is a constrained MDP (CMDP) problem, which is a typical MDP problem with additional constraints. Solving such a CMDP problem with uncertainty is challenging. On the one hand, typical MDP problems are well-investigated, which can be solved by iterative methods by finding a deterministic policy, such as the policy iteration and the value iteration~\cite{9177293}.
However, these methods for MDP cannot cope with the CMDP problem since constraints and the objective cannot be optimized simultaneously.
On the other hand, although CMDP problems with the known transition probability can be solved simply via a linear programming method, the linear programming method cannot address the CMDP problem with uncertainty, since transition probability $P(H_{t+1}|H_t)$ is unknown due to the uncertainty of the arrived task number.}

\section{Deep risk-sensitive RL Algorithm}
In this section, we first introduce the preliminary of RL methods. Afterward, by tailoring the typical RL methods, we propose the deep risk-sensitive RL algorithm to address problem P2. Finally, we present the details of DNN-based implementation of the proposed algorithm.

\subsection{Preliminary}
\bl{In problem P2, since the objective is to find policy $\bm \pi$ that chooses appropriate actions at different states to minimize the long-term cost (delay), which consists of the immediate cost (generated in the current epoch) and the future cost (generated in the following epochs) for each state-action pair. Because the future cost is related to both the current scheduling action and the actions in the following epochs, it is challenging to model the relationship between the current action and the future cost, particularly in the case with unknown state transition probability. Therefore, the discounted cost model is designed to balance the immediate cost and the future cost for each state-action pair, which is calculated as $\sum_{t=0}^{\infty}{\varsigma^t C(\bm s_t, \bm a_t)}$~\cite{puterman2014markov}. Note that the discount factor, denoted by $\varsigma \in \left[0, 1\right]$ is to prevent the long-term cost from going to negative infinity.}

\bl{Then, to measure the long-term cost starting from state $\bm s$ under policy $\bm \pi$, a \emph{value function} is defined to determine the value of expected long-term discounted cost when the system is at state $\bm s$. Denote by $V_{ \bm \pi}(\bm s)$ the value function, which is given by:}
\begin{equation}\label{eq17}
V_{ \bm \pi}(\bm s) = \mathbb E \left[ \sum_{t=0}^{\infty}{\varsigma^t C(\bm s_t, \bm a_t)} |\bm \pi, \bm s_0 = \bm s\right].
\end{equation}
\bl{Based on \ref{eq17}, a \emph{Q-value function} is defined to further evaluate state-action pairs, which is denoted by $Q_{ \bm \pi}(\bm s_t, \bm a_t)$. Such the Q-value function measures the expected long-term discounted cost that the system may get from being at state $\bm s$, following policy $\bm \pi$ and choosing action $\bm a$, which is given by:
\begin{equation}
Q_{ \bm \pi}(\bm s_t, \bm a_t) = C(\bm s_t, \bm a_t) + \sum_{\bm s_{t+1}}{\varsigma P(\bm s_{t+1}|\bm s_t, \bm a_t)V_{\bm \pi}(\bm s_{t+1})}.
\end{equation} 
With the objective of the cost minimization, we choose the minimum Q-value as the optimal Q-value, which is denoted by $Q^{*}_{\bm \pi}(\bm s, \bm a) = \min_{\bm \pi} Q_{\bm \pi}(\bm s, \bm a)$.}

\bl{Generally, due to the unknown state transition probability, the basic idea behind model-free RL methods is temporal difference (TD) learning, i.e., the current approximation of Q-value function (which might not be accurate) can be leveraged to update the estimated value for the following states~\cite{mnih2015human}. The mechanism of the RL methods allows the UAV to iteratively update and approximate the Q-value function and then choose actions based on the approximated Q-value function. Therefore, RL methods can learn online and interact with the environment simultaneously, which is suitable for the considered case with unknown task arrival. Denote by $\bm a^{*} = \ \arg \min_{\bm a_t \in \bm A} Q_{\bm \pi}(\bm s_t, \bm a_t)$ the greedy action which acquires the optimal Q-value. The Q-value can be updated based on the following TD backup equation:    
\begin{equation}
Q_{\bm \pi}^{\prime}(\bm s_t, \bm a_t) = Q_{\bm \pi}(\bm s_t, \bm a_t) + \eta \left[ C(\bm s_t, \bm a_t) + \varsigma Q_{\bm \pi}(\bm s_{t+1}, \bm a^{*}) \right],
\end{equation}
where the learning rate denoted by $\eta$ is to determine how much newly acquired cost should be accepted to adjust the evaluation of Q-value function. Note that $0 <\eta <1$ is a constant value in the learning process. The convergence of such RL methods based on Q-value iteration has been proved, i.e., the Q-values converge to the optimal Q-values~\cite{puterman2014markov}.}

\bl{Conventional RL methods update Q-values based on a Q-value table, i.e., all state-action pairs are listed in a table, and each pair is updated iteratively and independently. However, tabular methods require a large memory to store all state-action pairs, which increases exponentially with the state and action space~\cite{mnih2015human}. Due to the curse of dimensionality in the considered scenario (e.g., a large number of UAV locations, the large size of the computing queue backlog), conventional tabular RL methods cannot be applied practically.} To deal with the aforementioned problem, instead of tabular methods, DNN is adopted to approximate Q-value function~\cite{specht1991general}. Let $\vartheta$ be the parameters of DNN, which includes neural network weights and biases. Denote by $Q_{\bm \pi}(\bm s_t, \bm a_t; \vartheta)$ the DNN-based Q-value function, which is updated by minimizing the following loss function:
\begin{equation}\label{eq12}
 L(\vartheta) = | C(\bm s_t, \bm a_t) + \varsigma Q_{\bm \pi}(\bm s_{t+1}, \bm a^{*}; \vartheta) - Q_{\bm \pi}(\bm s_t, \bm a_t; \vartheta) |^2,
\end{equation}
where $ L(\vartheta)$ is named as the TD error. \bl{Similar to tabular RL methods, DNN-based RL methods can also allow the UAV to iteratively update the DNN-based Q-value function and then choose actions based on the approximated DNN-based Q-value function in an online manner.}

\subsection{The Deep Risk-Sensitive RL Algorithm Design}
\bl{In problem~P2, apart from the objective of cost minimization, there is an extra constraint of energy capacity that needs to be satisfied. However, since the energy consumption is not a component of the cost function, conventional RL methods mentioned above cannot satisfy the constraint in problem~P2. Therefore, we propose a deep risk-sensitive RL algorithm to deal with the CMDP problem. Specifically, in addition to the cost function, an extra risk function is defined to capture whether the UAV energy consumption in the current epoch violates the UAV energy capacity constraint, and then a corresponding Q-value function is defined to evaluate the value of risk. Therefore, the algorithm has two Q-value functions, i.e., one Q-value function to evaluate the cost and the other Q-value function to evaluate the risk. Afterward, the proposed deep risk-sensitive RL algorithm updates two different Q-value functions independently and chooses the action based on the sum of two Q-value functions.}

\bl{Define the set of error states as $\varPhi \subseteq \bm S$. An error state $\bm s_t \in \varPhi $ represents the energy consumption of the UAV in epoch $t$ exceeds the UAV energy capacity, i.e., $E_t > \varepsilon t$. Then, to measure how much consumed energy that exceeds the UAV energy capacity when the system is at state $\bm s$ choosing action $\bm a$, we denote the risk function by $R(\bm s_{t}, \bm a_t)$, which is given by:
 \begin{equation}
 R(\bm s_{t}, \bm a_t) = \left\{
 \begin{aligned}
 & | E_t - \varepsilon t |, &&\;\text{if} \,\, \bm s_t \in \varPhi\\
 & 0, &&\;\text{otherwise}.\\
 \end{aligned}
 \right.
 \end{equation}
The value of risk that are at a non-error state is zero, and the value of risk at an error state is equivalent to the exceeding part of the energy consumption. Consequently, if the current state of the system is an error state, the following states will also be error states with the increased value of risk. To satisfy the UAV energy capacity constraint in problem~P2, the value of risk at each state should be zero. Thus, we transform the goal that keeps the energy consumption below the energy capacity into the goal that minimize the risk. Note that the risk minimization is not equivalent to energy consumption minimization since the energy consumption minimization is not the objective of this problem.} 

Similar to the aforementioned cost minimization, the risk minimization can be achieved by using another Q-value function, which is operated separately. 
Based on the discounted risk, we define the expected long-term discounted risk as the value function $\bar V_{ \bm \pi}(\bm s)$ of state $\bm s$ under policy $\bm \pi$, which is given by:
\begin{equation}
\bar V_{ \bm \pi}(\bm s) = \mathbb E \left[ \sum_{t=0}^{\infty}{\bar \varsigma^t R(\bm s_t, \bm a_t)} |\bm \pi, \bm s_0 = \bm s\right],
\end{equation}
where $\bar \varsigma$ is the discount factor for the discounted risk.
Then, to measure the expected long-term discounted risk that the UAV may get from being at state $\bm s$, following policy $\bm \pi$ and choosing action $\bm a$, the corresponding Q-value function, $\bar Q_{\bm \pi}(\bm s_t, \bm a_t)$, is defined as follows:
\begin{equation}
\bar Q_{ \bm \pi}(\bm s_t, \bm a_t) = R(\bm s_t, \bm a_t) + \sum_{\bm s_{t+1}}{ \bar \varsigma P(\bm s_{t+1}|\bm s_t, \bm a_t) \bar V_{\bm \pi}(\bm s_{t+1})}.
\end{equation}
Based on the TD learning, the Q-value function of risk can also be estimated based on the following equation:
\begin{equation}
\bar Q_{\bm \pi}^{\prime}(\bm s_t, \bm a_t) = \bar Q_{\bm \pi}(\bm s_t, \bm a_t) + \bar \eta \left[ R(\bm s_t, \bm a_t) + \bar \varsigma \bar Q_{\bm \pi}(\bm s_{t+1}, \bar{\bm{a}}^{*} ) \right],
\end{equation}
where $\bar \eta$ is the learning rate for the risk minimization, and greedy action $\bar{\bm{a}}^{*} = \text{arg} \min_{\bm a_t \in \bm A} \bar Q_{\bm \pi}(\bm s_t, \bm a_t)$ is adopted to acquire the optimal Q-value. \bl{As $\bar{\bm{a}}^{*} $ and $\bm a^{*}$ are two different greedy actions based on different goals, i.e., cost minimization and risk minimization, the chosen actions may not be the same at each state. However, only one action can be selected when each state is reached.} Thus, we need to design a new Q-value function to combine two goals. which is given by:
\begin{equation}
Q_{\bm \pi}^{\delta}(\bm s_t, \bm a_t) = Q_{\bm \pi}(\bm s_t, \bm a_t) + \delta \bar Q_{\bm \pi}(\bm s_t, \bm a_t),
\end{equation}
where $\delta$ is a weight parameter to balance two different goals. If $\delta$ is fixed, $Q^{\delta}$ forms a standard Q-value function of state-action pair with respect to the new reward $C + \delta R$, which is same as the Q-value function in typical RL methods~\cite{geibel2005risk}~\cite{xiao2019reinforcement}. Specifically, when $\delta = 0$, $Q^{\delta} = Q$, the minimization of the weighted sum of the cost and the risk leads to the optimal policy for cost minimization, which is same as the cost minimization without constraints. When $\delta$ tends to infinity, the minimization of the weighted sum of the cost and the risk leads to the optimal policy for the risk minimization.  As the adaption of $\delta$ provides a method to find the space of feasible polices, $\delta$ can be adjusted to produce the optimal policy to minimize the cost while satisfying the constraint. Therefore, there exists the optimal deterministic policy for the designed new Q-value function, and the convergence of proposed deep risk-sensitive RL algorithm can be guaranteed if discount factors $\varsigma$ and $\bar \varsigma$ are equivalent \cite{mnih2015human}. 

\begin{algorithm}[t]\label{Alg:DOTS}
 \caption{Deep Risk-Sensitive RL Algorithm}
 \LinesNumbered
 \SetKwInOut{Input}{Input}
\textbf{Initialize:} $\varepsilon$, replay memory $D$; $\vartheta$, $\vartheta'$, $\bar \vartheta$, $\bar \vartheta'$; state ${\bm s}_0$; step size $\Delta$; $\delta$;\\
\For{$k=1,2,3,\cdots,K$}
{   
    \For{$t=1,2,3,\cdots,T$}
     {
      Choose $\bm a_t$: select a random action with probability $\epsilon$, or select $\arg\min\limits_{\bm a}{\left[Q(\bm s_t,\bm a;\vartheta) + \delta \bar Q(\bm s_t,\bm a; \bar \vartheta)\right]}$ with probability $1 - \epsilon$;\\
     Perform action $\bm a_t$ and observe cost $C_t$, risk $R_t$ and next state $\bm s_{t+1}$;\\
     Store transition $\left(\bm s_t,\bm a_t, C_t, R_t, \bm s_{t+1}\right)$ in $D$;\\
     Sample \!random mini-batch of \!transitions\!  $\left(\!\bm s_j,\!\bm a_j,\!C_j,\!R_j,\!\bm s_{j+1}\!\right)$ from $D$;\\
     Set $y_j = C_j + \varsigma \min\limits_{\bm a'}(Q'(\bm s_{j+1},{\bm a'};\vartheta'))$;\\
      Set $\bar y_j = R_j + \bar \varsigma \min\limits_{\bm a'}(\bar Q'(\bm s_{j+1},{\bm a'};\bar \vartheta'))$;\\
      Perform a gradient descent step on $\mathbb{E}_{\left(\bm s_j,\bm a_j, C_j, R_j, \bm s_{j+1}\right) \sim U(D)}{\left[(y_j-Q(\bm s_j,{\bm a_j};\vartheta))^2 \right]}$ with respect to $\vartheta$;\\
      Perform a gradient descent step on $\mathbb{E}_{\left(\bm s_j,\bm a_j, C_j, R_j, \bm s_{j+1}\right) \sim U(D)}{\left[(\bar y_j- \bar Q(\bm s_j,{\bm a_j};\bar \vartheta))^2 \right]}$ with respect to $\bar \vartheta$;\\
      Set $\vartheta'$ = $\vartheta$, and $\bar \vartheta'$ = $\bar \vartheta$;\\
     }

\eIf{$\frac{E_T}{T} > \varepsilon$}
{$\delta \leftarrow \delta + \Delta$\;}
{$\delta \leftarrow \delta - \Delta$;}

 }
\textbf{Output:} DNN models with parameters $\vartheta$ and $\bar \vartheta$, and weight parameter $\delta$
\end{algorithm}

\bl{Due to the curse of dimensionality, we adopt DNN to approximate the Q-value function of risk as the approximation of the DNN-based Q-value function of cost. Denote by $\bar Q_{\bm \pi}(\bm s_t, \bm a_t)$ the DNN-based Q-value function of risk, where $\bar \vartheta$ is the parameter of the corresponding neural network. The update of DNN-based Q-value function of risk is the same as that of Q-value function of cost in (\ref{eq12}). As shown in Algorithm~1, we propose a two-cycle algorithm to minimize the cost while minimizing the risk, i.e., learn the appropriate parameters of DNNs in the inner cycle, and search the appropriate weight parameter to balance two goals in the outer cycle. The former is shown from line~4 to line~20, and each inner cycle is named as an \emph{iteration}. In one iteration, the DNN parameters of $ Q_{\bm \pi}(\bm s_t, \bm a_t; \vartheta)$ and $\bar Q_{\bm \pi}(\bm s_t, \bm a_t; \bar \vartheta)$ are updated separately and iteratively. The searching in the outer cycle is shown from line~3 to line~21. Each outer cycle is named as an \emph{episode}. In each outer cycle, the optimal weight parameter $\delta$ is updated according to the energy consumption, which is shown from line~16 to line~20. Based on whether the energy consumption in the current episode satisfies the constraint, weight parameter $\delta$ is increased or decreased with a fixed step size denoted by $\Delta$. The partial detail of Algorithm~1 is introduced in the next subsection.}

\subsection{DNN-based Implementation}
\begin{figure*}
	\centering
  	\includegraphics[width = 2\columnwidth]{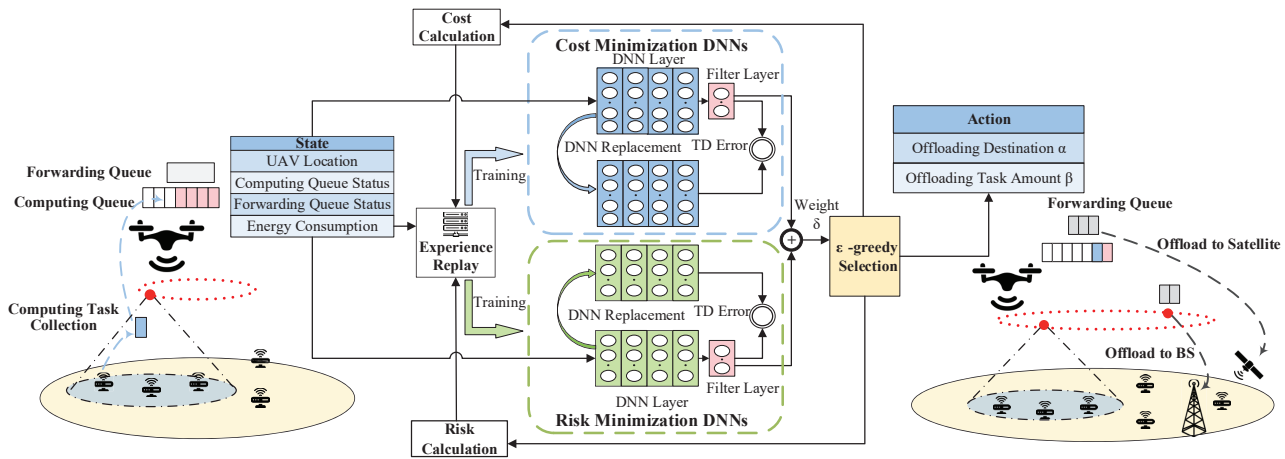}
  	\caption{\small An overview of the deep RL-based DOTS scheme.}\label{model}
\end{figure*}
Instead of constructing space-costly Q-tables in conventional RL methods, we implement the proposed algorithm by approximating the Q-value function via DNNs. However, directly replacing the Q-table by a DNN model meets several challenges, e.g., unavailable actions at each state cannot be deleted simply by DNNs due to the ``black-box'' characteristic of DNN. Therefore, we should design the DNN model to fit the proposed algorithm, the details of which are introduced as follows. As shown in Fig.~\ref{model}, four significant modules are introduced, i.e., DNN replacement, filter layer design, experience replay, and $\epsilon$-greedy selection. 

\subsubsection{DNN Replacement}
For a more stable training, we adopt two DNNs to estimate a Q-value function, i.e., one for a target network and the other for a prediction network. The target network has the same DNN architecture as the prediction network but with frozen parameters. For every certain number of iterations, the parameters from the prediction network are copied to the target network, and this procedure is called DNN replacement. Since the TD error is used as the loss function in DNN backpropagation to approximate Q-value function by DNNs, the backpropagation requires the output gradient of DNN with respect to weights for input epoch $t$, and this gradient needs to be saved until we have the new TD error at epoch $t+1$. Thus, there always exists a predicted value for epoch $t+1$ when we compute gradient at epoch $t$. If we use the same DNN to calculate the predicted value (e.g., $Q_{\bm \pi}(\bm s_t, \bm a_t; \vartheta)$) and the target value (e.g., $C(\bm s_t, \bm a_t) + \varsigma Q_{\bm \pi}(\bm s_{t+1}, \bm a^{*}; \vartheta)$), the DNN can become destabilized in the feedback loops between the target value and the predicted value \cite{mnih2015human}. Considering cost minimization and risk minimization are independent, we leverage two DNNs to approximate Q-value function of cost, and another two DNNs to estimate Q-value function of risk, which are shown in Fig.~\ref{model}.

\subsubsection{Filter Layer Design}
We adopt a filter layer to exclude the outputs of unavailable actions. In the considered problem, the available action set at different states is different. For example, the UAV can only offload tasks to the nearby BSs, and thus available action set $\mathcal L_t$ changes with the location of UAV $l_t$. However, since the output size of a fully connected layer in DNN is fixed, the number of Q-value outputs from DNN cannot be changed according to the various number of actions in the available set. As a result, the unavailable actions are included in the DNN-based approximation of Q-value, which is incorrect. Furthermore, constraint~\eqref{cons1} needs to be guaranteed and requires the various available action set at different states. Thus, we adopt a binary coding in the filter layer, which can select available action depending on the current state. Then, to exclude unavailable actions, the Q-value of these actions can be increased (i.e., add a constant to the original Q-value, which is a hyper-parameter depending on the magnitude of Q-values). These actions are excluded since only the minimal Q-value is selected to feed into the loss function. As shown in Fig.~\ref{model}, a filter layer is added to help the target network exclude invalid actions and output real Q-values.
 
\subsubsection{Experience Replay}
Considering the high correlation between continuous states in this scenario (e.g., cumulative energy consumption $E_t$ is highly correlated with $E_{t+1}$ due to the accumulative sum), DNN can be easily over-fitting if high correlation data is fed. Furthermore, the DNN is required to not only learn from current interaction with the environment but also a more varied array of past experiences (e.g., past task arrival pattern). To this end, experience replay is utilized to store experiences including state transitions, costs, risks, and actions, which are necessary to perform the proposed deep risk-sensitive RL. \bl{As shown in Fig.~\ref{model}, the replay memory, denoted by $D$, is used to store experience, and mini-batches of experiences are fed to train DNNs. In Algorithm~1, mini-batches of experience $\left(\bm s_j,\bm a_j, C_j, R_j, \bm s_{j+1}\right) \sim U(D)$ are uniformly draw at random from the replay memory to update DNNs.} This technique has the following merits: 1) reducing the correlation among experiences in updating DNNs, 2) reusing the previous state transitions to avoid catastrophic forgetting, and 3) increasing learning efficiency with mini-batches and learning stability.  

\subsubsection{$\epsilon$-Greedy Selection}
To learn how to react to all possible states in the environment, it must be exposed to as many as possible states. The UAV needs to explore different energy consumption and the number of tasks in the buffer. \bl{However, the UAV needs to exploit the exposed experiences to learn a decent task scheduling policy, which conflicts the experience exploration. Thus, the proposed learning policy should deal with such an exploration and exploitation trade-off.} To deal with this problem, the $\epsilon$-greedy selection approach is leveraged to balance the trade-off. The UAV selects the action based on approximated Q-value function most of the time, but occasionally chooses the action randomly. In the realization of Algorithm~1, parameter $\epsilon$ is an adjustable parameter which determines the probability of taking a random action, rather than the action based on the Q-value function. 

\section{Performance Evaluation}
In this section, extensive simulations are carried out to evaluate the proposed deep RL-based  DOTS scheme. Specifically, we first elaborate on the simulation settings, and benchmark strategies. Afterward, the overall performance evaluation of the proposed scheme is conducted.

\subsection{Simulation Settings}
In the experiments, locations of IoT devices follow a uniform distribution~\cite{8672604}. The computing task arrival is set to follow a Poisson distribution with arrival rate $\mu$, which is unknown \emph{a priori} for the UAV~\cite{chau2019delay}. Referring to well-studied UAV trajectory design algorithm \cite{zeng2017energy}, a UAV is dispatched. The UAV flies along with main areas of IoT devices, which can be more effective to accommodate the IoT service demand. The UAV trajectory is generated by the VISSIM which is a simulation tool in transportation research~\cite{zhang2018air}. The altitude of the UAV is set to 10\,m, and the size of computing queue $\rho$ is set to 20. Additionally, by adopting the pathloss (in dB) model of UAV communication in~\cite{8672604}, the pathloss of UAV-BS links is given by:
\begin{equation}\label{eq8}
    PL\left(x, \theta \right) =10 A_0 \log\left(x\right) + B_0\left(\theta  - \theta_0\right)e^{\frac{\theta_0 - \theta}{C_0}} + \eta_0,
\end{equation}
where $x$ represents the distance between the UAV and a BS, and $\theta$ is the corresponding vertical angle. Both $x$ and $\theta$ can be obtained based on UAV location $l_t$ and the BS location. Due to the mobility of the deployed UAV, $x$ and $\theta$ vary over different locations. 
Parameters $A_0$, $\theta_0$, $B_0$, $C_0$ and $\eta_0$ in (\ref{eq8}) are configured as 3.04, -3.61, -23.29, 4.14, and 20.7, respectively~\cite{8672604}.
Meanwhile, the LEO satellite connection is always available for the UAV. The Weibull-based channel model is adopted to model the rain attenuation of UAV-satellite links \cite{kanellopoulos2014channel}. Other simulation parameters are listed in Table I.

\begin{table}[t]
\normalsize
\centering
\captionsetup{justification=centering,singlelinecheck=false}
\caption{Simulation Parameters}\label{table1}
\begin{tabular}{c|c|c|c}
\hline\hline
 Parameter & Value & Parameter & Value\\
 \hline\hline
 $N$ & 5 & $f_\text{U}$ & 1\,Gigacycle/s \\
 \hline
 $\phi$ & 5\,MB  & $f_0$ & 5\,Gigacycle/s \\
 \hline
 $\gamma$  & 25\,cycles/bit  & $f_n, n\not=0$ & 10\,Gigacycle/s\\
 \hline
 $W_\text{B}$ & 3\,MHz & $N_0$ & -174\,dBm/Hz \\
 \hline
 $W_\text{S}$ & 2\,MHz & $P_\textrm{B}$ & 1.6\,W  \\
 \hline
 $P_\text{S}$ & 5\,W & $\xi$ & $10^{-28}$ \\
 \hline
 $d_\text{S}$ & 6.44\,ms & $\beta^\text{max}$ & 7\\
 \hline
\end{tabular}
\end{table}

The proposed DNN-based scheme is implemented via Python 3.7 and Tensorflow open-source machine learning library \cite{abadi2016tensorflow}. The training of DNNs is conducted with a NVDIA 1660\,Ti GPU. The DNN of cost minimization includes four fully-connected hidden layers with (256, 128, 128, 64) neurons, and the DNN of risk minimization includes four fully-connected hidden layers with (512, 256, 128, 128) neurons, respectively. ReLU function is adopted as the activation function to realize nonlinear approximation after the fully connected layers. Additionally, L2 regularization is used to reduce the possibility of DNN over-fitting. Meanwhile, Adam optimizer is adopted in the DNN training. In each episode, the behavior policy during training is $\epsilon$-greedy with $\epsilon$ increases linearly from 0 to 0.9995 over 35,000 iterations. 

Benchmark schemes adopted in this computing task scheduling problem are introduced below:
\subsubsection{Random probabilistic configuration (RPC)}
In this scheme, the random policy is adopted, which means that actions are selected randomly in different states. All available actions are selected with the same probability.
\subsubsection{Sampling-based probabilistic configuration (SPC)}
In this scheme, the probability of available actions on each state is fixed. Based on a large number of historical sampling experiments, the probability of different actions is configured to meet the UAV energy capacity.  

\subsection{Simulation Results}
\bl{We show the simulation results of our proposed algorithm from two parts. Firstly, we evaluate the convergence performance of the proposed deep RL-based DOTS scheme. Secondly, we compare the performance of the proposed deep RL-based DOTS scheme with other benchmark schemes.}

\begin{figure*}
	\centering 
	\subfigure[] { \label{fig7} 
	\includegraphics[width=0.95\columnwidth]{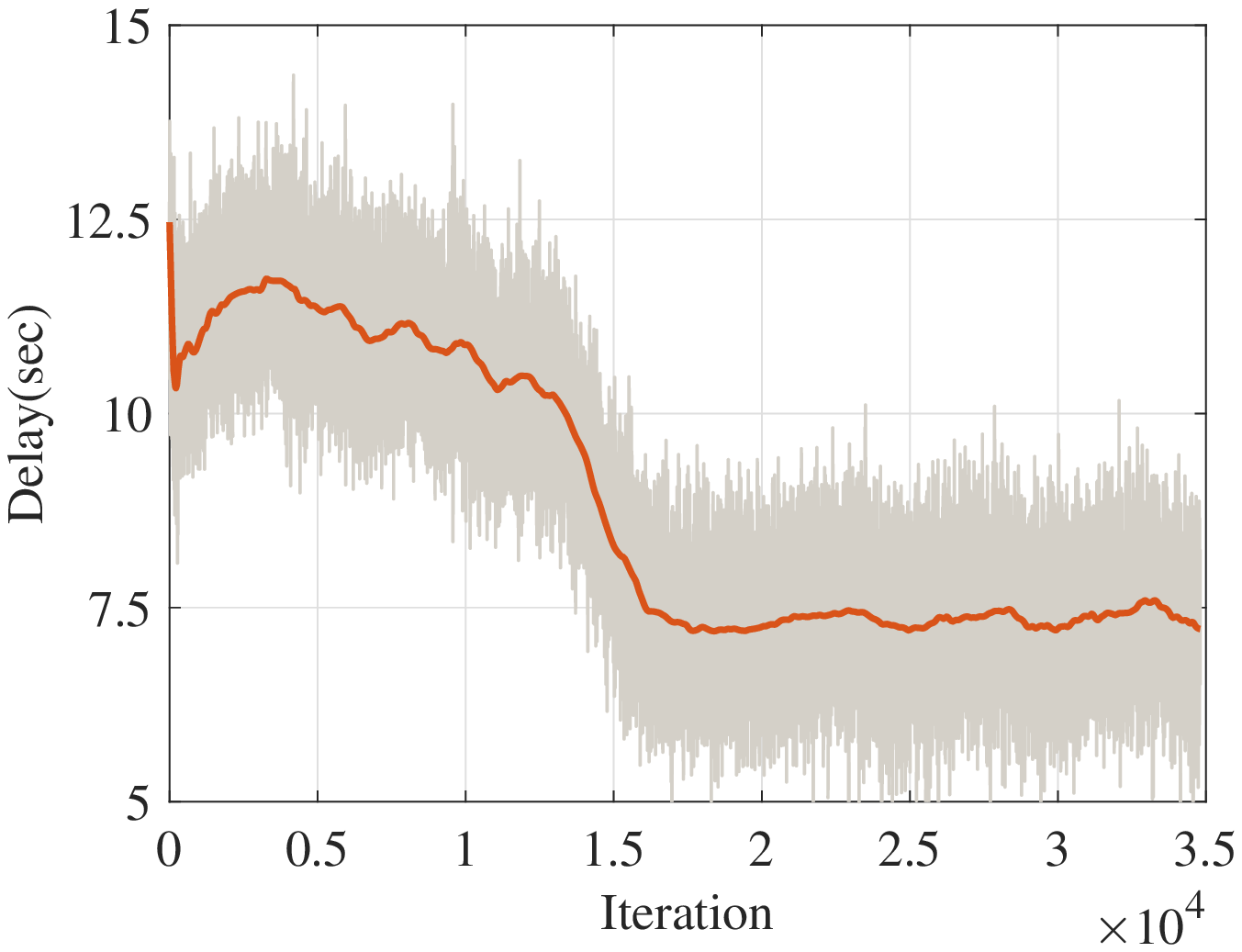} 
	} 
	\subfigure[] { \label{fig8} 
	\includegraphics[width=0.95\columnwidth]{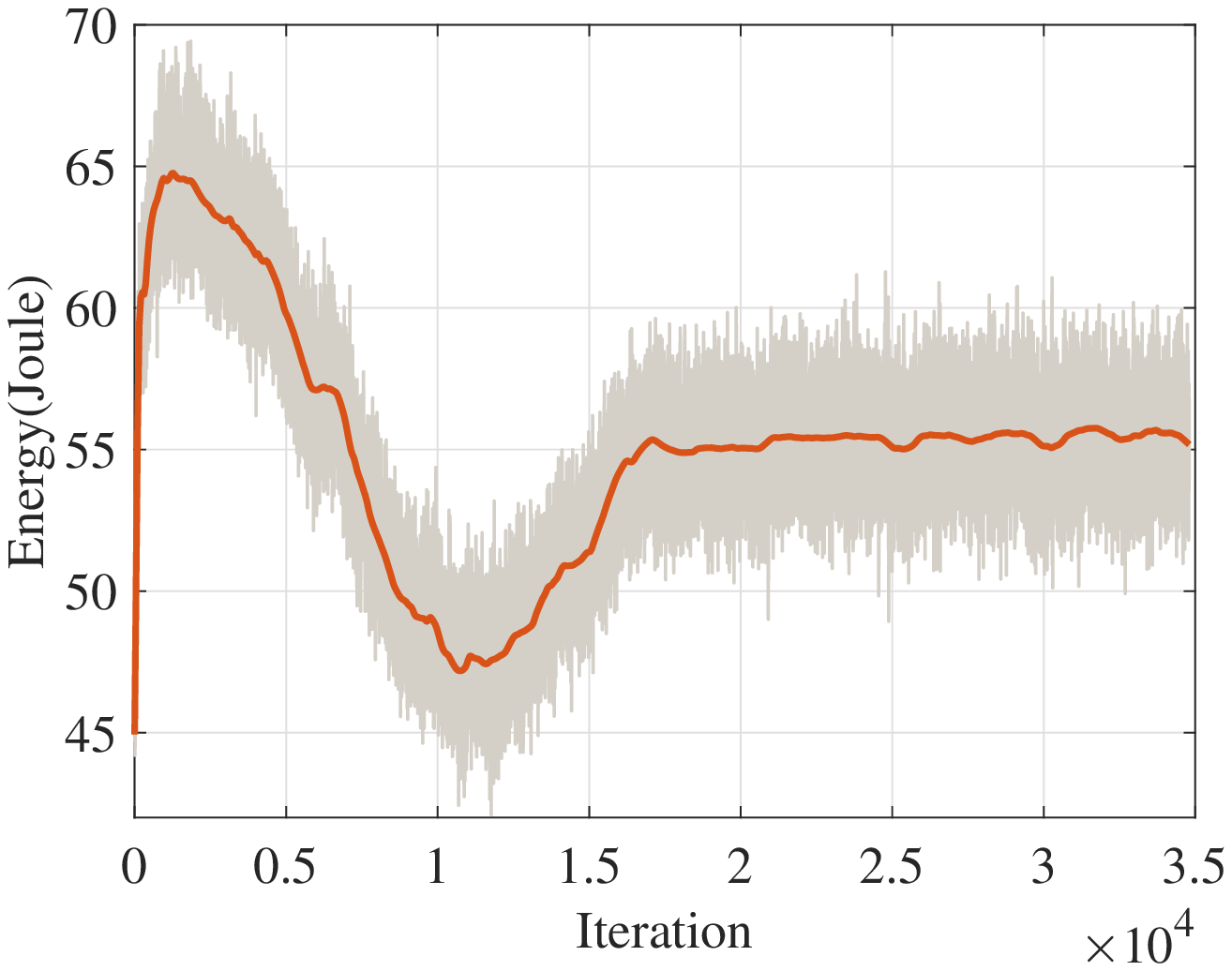} 
	} 
	\caption{\small Convergence performance of the proposed deep RL-based DOTS scheme in one episode.}\label{4}

\end{figure*} 
\subsubsection{Convergence Performance}
\bl{The convergence performance of the two-cycle structure of the proposed algorithm is shown in this subsection, i.e., the convergence performance of the inner cycle in Fig.~4 and that of the outer cycle in Fig.~5.} 

\bl{Fig.~4(a) shows the convergence performance of the delay and the energy consumption in the inner cycle (in one episode), respectively, where the orange line is the moving average results of the previous 100 iterations. It can be seen that the delay converges after 16,000 iterations, when UAV energy capacity $\varepsilon$ is set to 55\,Joule. However, the convergence trends of delay and energy consumption vary differently due to the differentiated functions of the cost and the risk. Specifically, the delay performance gradually decreases and converges after around 16,000 iterations, while the energy consumption performance exhibits a turning point at around the 11,000\,th iteration. Compared to the simple policy of minimizing the risk, e.g., the UAV can offload fewer tasks to reduce energy consumption intuitively, the policy of minimizing the cost is related to both the task arrival and the policy of minimizing the risk. As a result, as shown in Fig.~4(b), from iteration 0 to iteration 11,000, the policy of minimizing the risk has been well learned, while the learning process of cost minimization is still ongoing as shown in Fig.~4(a). After 16,000 iterations, the policy of delay minimization is learned while the energy consumption is approximately equivalent to the UAV energy capacity.}

\begin{figure*} \centering 
	\subfigure[] { \label{fig1} 
	\includegraphics[width=0.95\columnwidth]{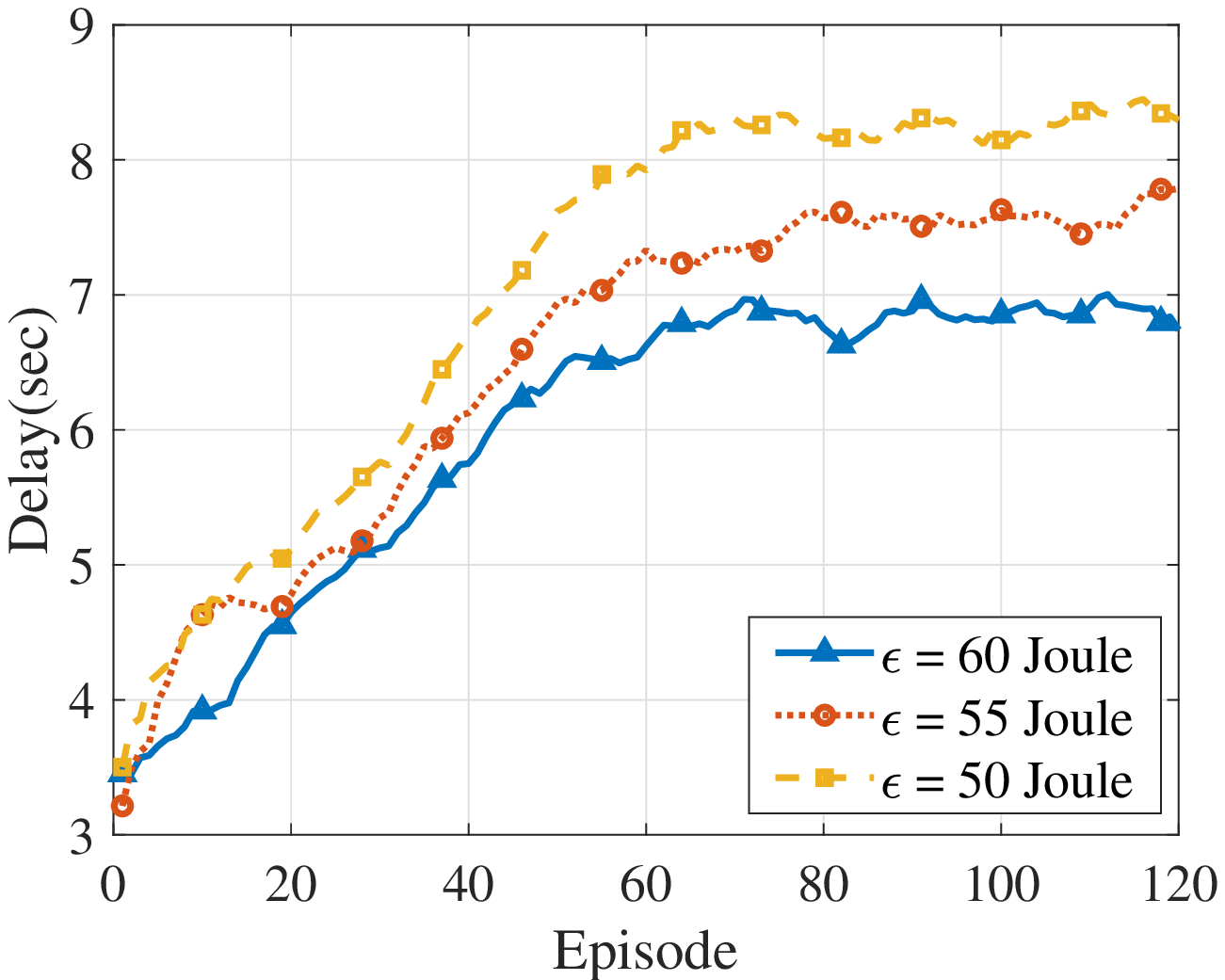} 
	} 
	\subfigure[] { \label{fig2} 
	\includegraphics[width=0.95\columnwidth]{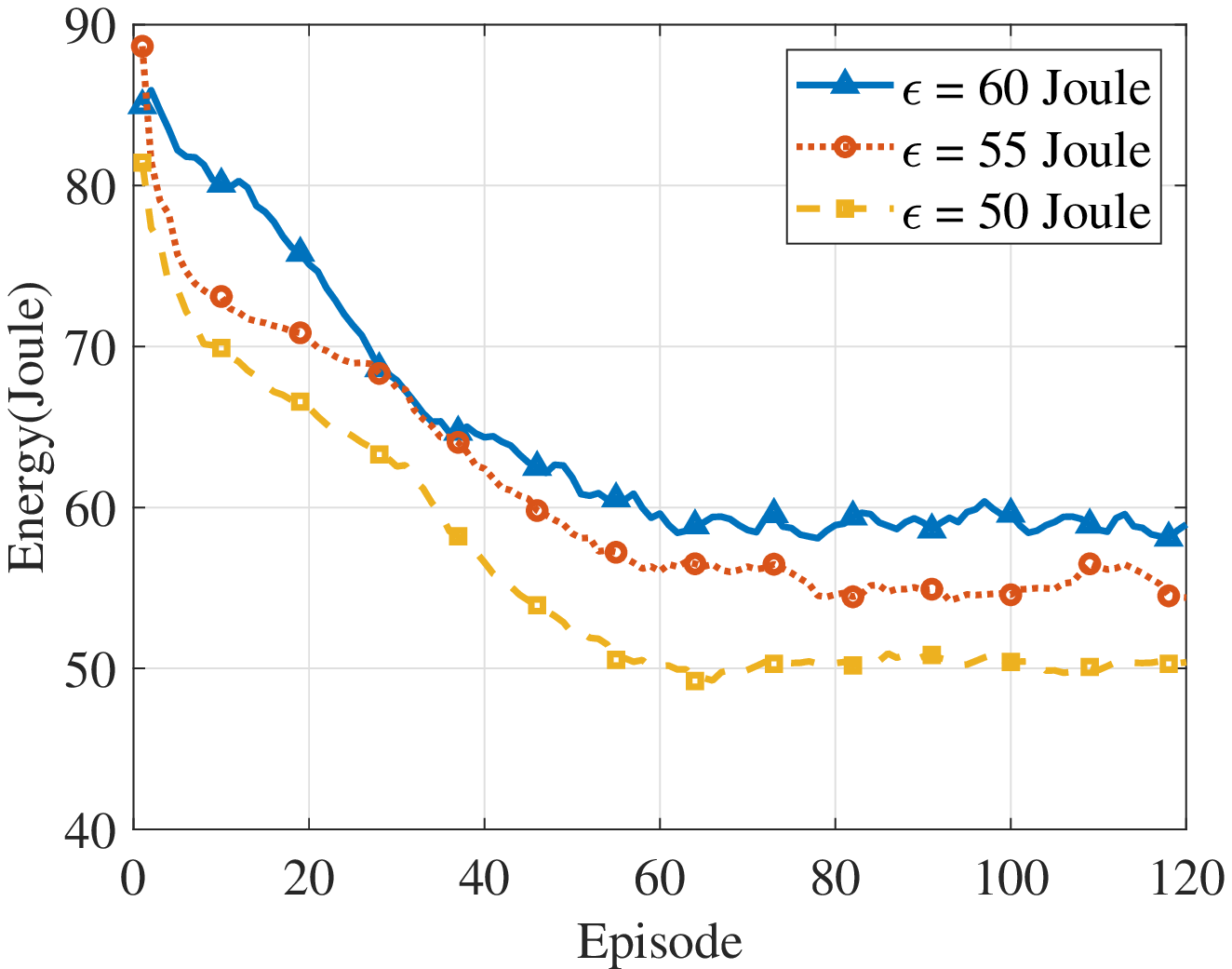} 
	} 
	\caption{\small Convergence performance of the proposed deep RL-based DOTS scheme.} 

\end{figure*} 

\bl{The convergence performance of delay and energy consumption in the outer cycle are shown in Fig.~\ref{fig1} and Fig.~\ref{fig2}, respectively. To evaluate the convergence performance of the proposed DOTS scheme, we adopt different values of the UAV energy capacity, i.e., 50\,Joule, 55\,Joule, and 60\,Joule. It can be seen that the average delay and average energy consumption converge after 70 episodes, where one episode consists of 35,000 iterations. Both the average delay and the average energy consumption oscillate at the beginning of the learning process due to the inaccurate weight parameter $\delta$, which takes time to approach to the optimal weight parameter. In Fig.~\ref{fig1}, we can observe that the average delay of the learned policy decreases as the increase of the UAV energy capacity of the UAV, which happens since more energy can be consumed by the UAV to offload more tasks to either the BS or the satellite. In Fig.~\ref{fig2}, the impact of energy consumption is shown on different energy consumption capacities. As expected, the energy consumption of different cases is approximately equivalent to the pre-set energy consumption capacities. Therefore, based on aforementioned convergence performance, the DOTS scheme can work well in scenarios with different energy consumption capacities.}

\begin{figure*} \centering 
	\subfigure[] { \label{fig5} 
	\includegraphics[width=0.95\columnwidth]{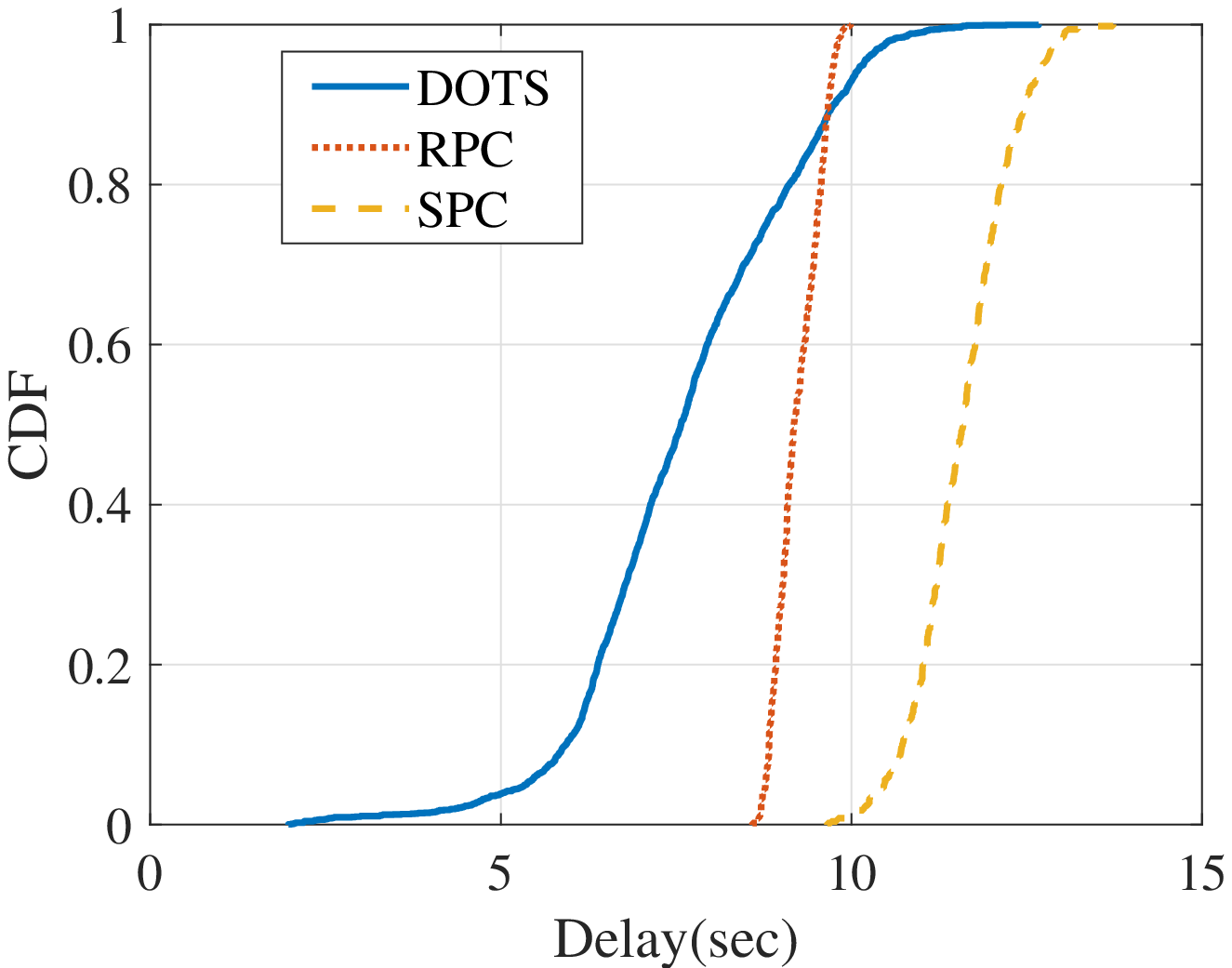} 
	} 
	\subfigure[] { \label{fig6} 
	\includegraphics[width=0.95\columnwidth]{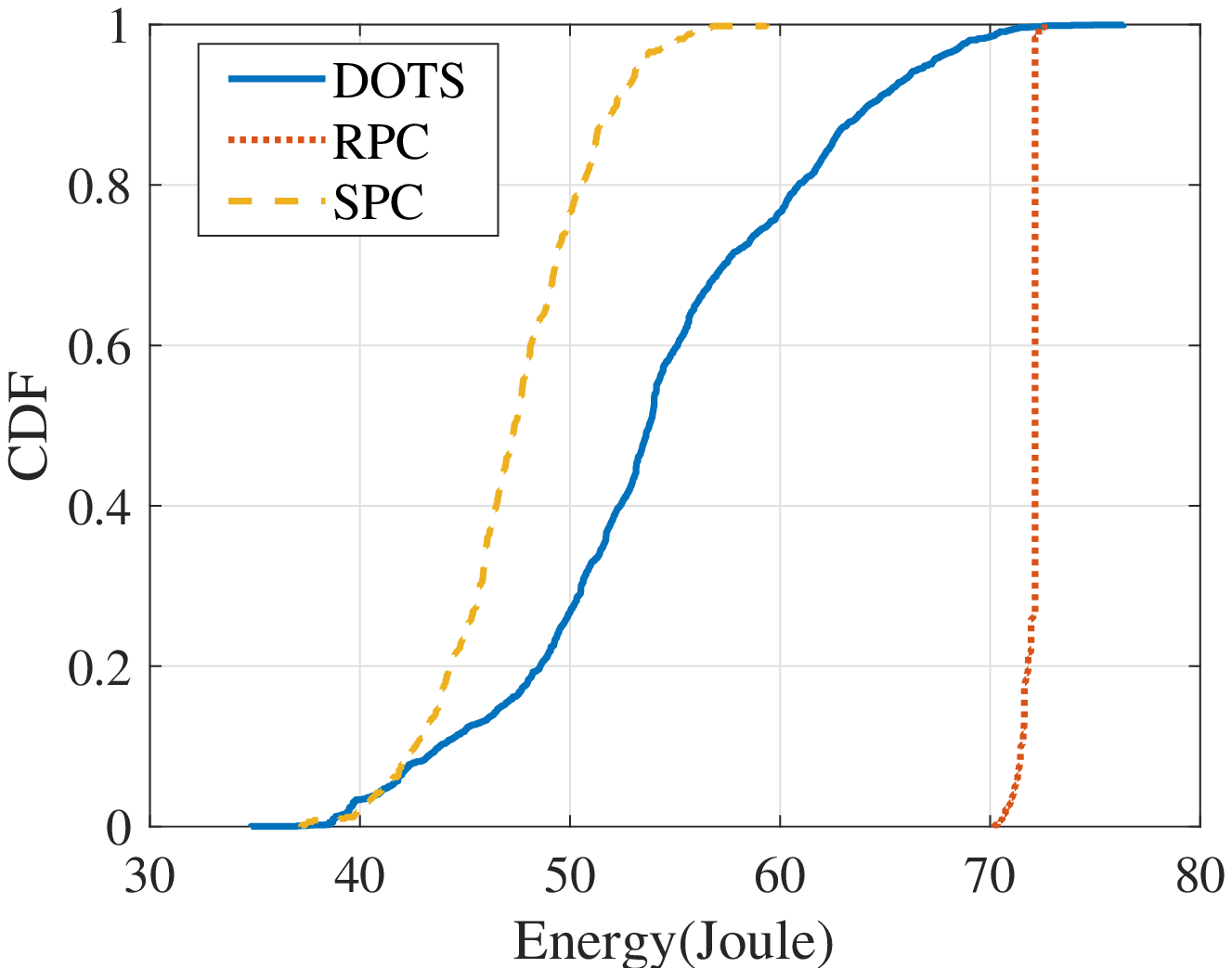} 
	} 
	\caption{\small CDFs of delay and energy consumption.} 

\end{figure*} 

\subsubsection{Performance Comparison}To compare DOTS with benchmark schemes, we plot cumulative distribution functions (CDFs) of delay and energy consumption in Fig. \ref{fig5} and Fig. \ref{fig6}, respectively. Note that average delay and energy consumption are calculated for the period that UAV flies back to the same destination along the same trajectory. 
Considering the dynamics of task arrival, we show the delay and energy consumption performance over 1,000 flights. 
We can see that DOTS is able to enhance the performance that the delay in 90\,$\%$ flights which is below 9 seconds. Meanwhile, 60\,$\%$ flights satisfy energy capacity of $\varepsilon$ = 55\,Joule. The RPC scheme cannot guarantee the UAV energy capacity constraint. Although the SPC scheme can satisfy the energy capacity, the delay of most flights is longer than 8.5 seconds. Therefore, the proposed DOTS scheme can work efficiently in different task arrival scenarios.

\begin{figure*} \centering 
	\subfigure[] { \label{fig9} 
	\includegraphics[width=0.95\columnwidth]{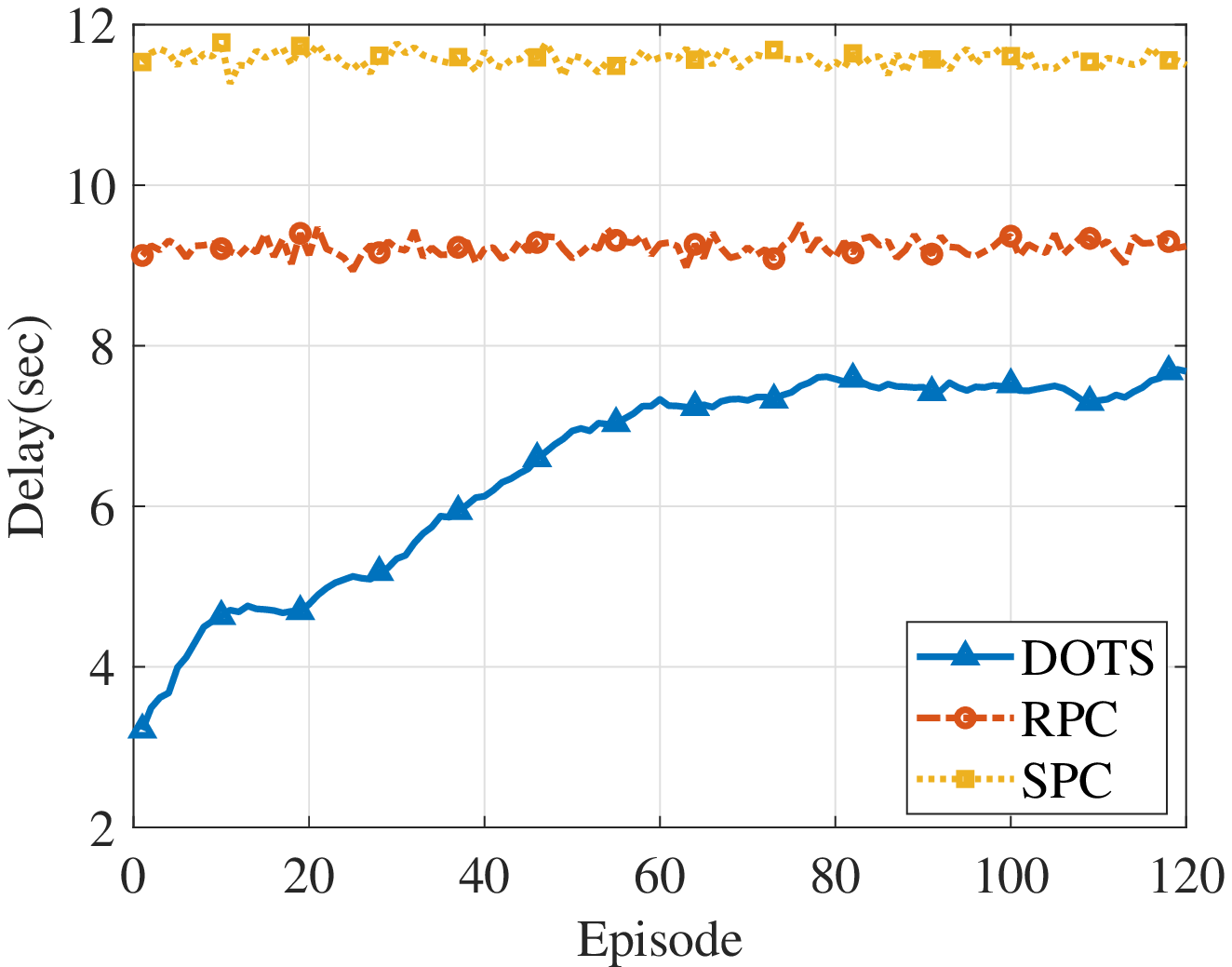} 
	} 
	\subfigure[] { \label{fig10} 
	\includegraphics[width=0.95\columnwidth]{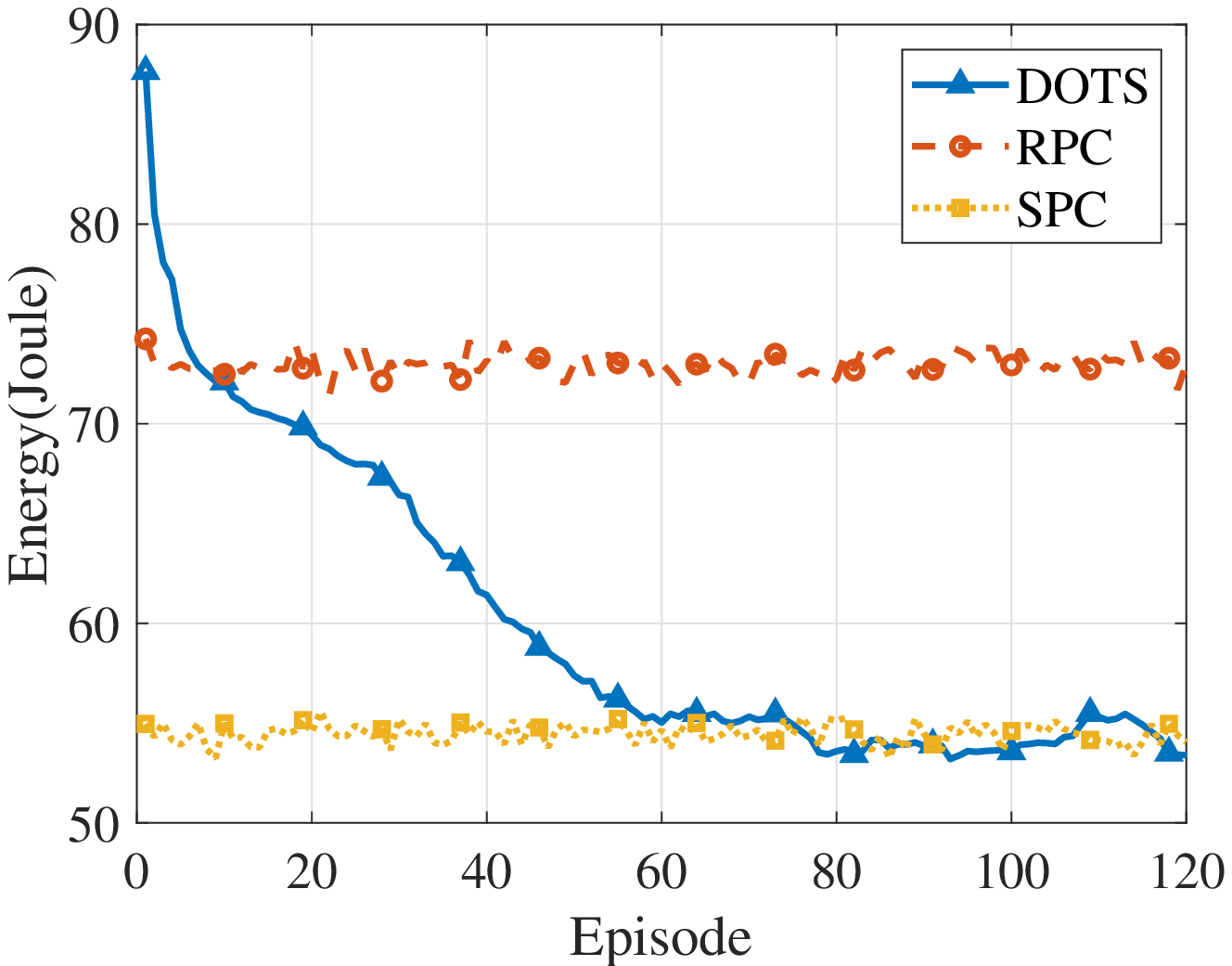} 
	} 
	\caption{\small Performance of delay and energy consumption.} 

\end{figure*} 

Figure \ref{fig9} and \ref{fig10} show the delay and energy consumption performance under DOTS, RPC, and SPC schemes, where the energy capacity is set to $\varepsilon = 55$\,Joule. In the simulation, we set the probability of offloading tasks in the SPC scheme to satisfy energy capacity 55\,Joule. It can be seen that the DOTS scheme and the SPC scheme are able to guarantee the UAV energy capacity constraint. However, the delay performance of the SPC scheme is worse than DOTS before 40 episodes, and the RPC scheme is always worse than the DOTS scheme. At the beginning of the learning process, the delay can be minimized, but the UAV energy capacity is exceeded. 
Due to the untuned weight $\delta$ at the beginning of the learning process, the goal of the policy is to minimize the cost. With the learning episode increasing, the policy of risk minimization can be found. Therefore, after 60 episodes, the delay-minimized policy is learned without exceeding the UAV energy capacity. Compared with the other two schemes, the proposed scheme has the lowest time-averaged task processing delay when the optimal policy has been learned.

\begin{figure}
	\centering
  	\includegraphics[width = 1\columnwidth]{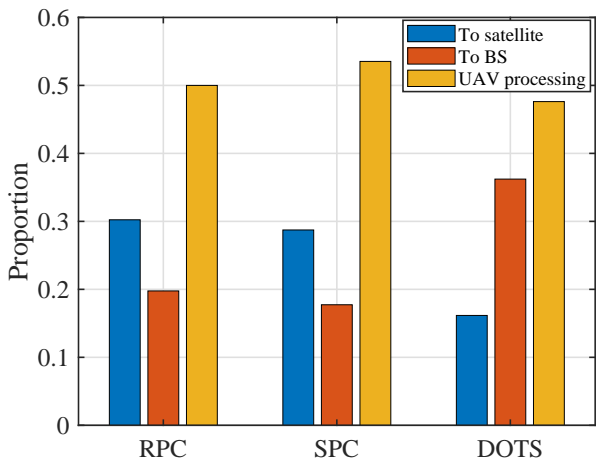}
	\caption{\small Offloading proportion under different policies for $\varepsilon$ = 55\,Joule.}\label{fig11}

\end{figure}

Figure \ref{fig11} shows the offloading proportion under different policies with $\varepsilon = 55$\,Joule. The action proportion of SPC and RPC schemes is similar, as both of them are based on probabilistic selection. However, RPC cannot guarantee the UAV energy capacity constraint. Although the SPC scheme can bound the energy consumption, SPC selects actions based on the historical experience, and thus it cannot learn to schedule proper number of tasks in different scenarios according to the future information. Particularly, the SPC scheme and the RPC scheme may offload the tasks at inappropriate states (e.g., low data rate), in which task offloading to other BSs or satellite should be suppressed and wait for more appropriate states (e.g., high data rate). Unlike the benchmark schemes, the proposed DOTS scheme can make the UAV offload a certain number of tasks to BSs when they are covered by BS, and offload to the satellite when it is out of the BS coverage. \bl{As offloading tasks to the satellite is an important complementary solution for offloading tasks to BSs, it effectively reduces the queuing delay when the UAV is out of the BS coverage. Therefore, the RL-based DOTS scheme can schedule the optimal number of tasks to BS or satellite according to the learned knowledge, such as the task arrival pattern.}

\section{Conclusion And Future Work}
In this paper, we have proposed a novel IoT computing task scheduling scheme named DOTS in SAGIN, where a UAV is dispatched to collect tasks from IoT devices and then make online scheduling decisions to process the tasks. Considering the limited UAV energy capacity and the dynamics of task arrival, we have formulated the online scheduling problem as a CMDP. With the objective of minimizing the long-term average delay without violating the constraint, we have designed the deep risk-sensitive RL algorithm to make online task scheduling decisions. Extensive simulation results have demonstrated that the deep RL-based DOTS scheme can significantly reduce the delay of processing IoT computing tasks while satisfying the UAV energy capacity constraint. The proposed scheme can provide low-latency IoT services and extend the service lifespan for massive IoT devices with limited power supply. In the future work, we will investigate the task scheduling strategy based on the cooperation of multiple UAVs in SAGIN.


\bibliography{ref}

\bibliographystyle{IEEEtran}

\end{document}